\documentclass[11pt,a4paper]{article}

\usepackage[utf8]{inputenc}
\usepackage[T1]{fontenc}
\usepackage{lmodern}
\usepackage{microtype}
\usepackage{amsmath,amssymb}
\usepackage{booktabs}
\usepackage{multirow}
\usepackage{array}
\usepackage{tabularx}
\usepackage{hyperref}
\usepackage{geometry}
\usepackage[table]{xcolor}
\usepackage{mdframed}
\usepackage{authblk}
\usepackage{parskip}
\usepackage{natbib}
\usepackage{graphicx}
\usepackage{caption}
\usepackage{float}
\usepackage[section]{placeins}  % \FloatBarrier at section boundaries
\usepackage{fontawesome5}
\usepackage{alltt}
\usepackage{enumitem}
\usepackage{listings}
\lstdefinestyle{promptstyle}{
  basicstyle=\footnotesize\ttfamily,
  breaklines=true,
  breakindent=0pt,
  breakautoindent=false,
  columns=fullflexible,
  keepspaces=true,
  showstringspaces=false,
  frame=single,
  rulecolor=\color{gray!45},
  backgroundcolor=\color{gray!6},
  xleftmargin=6pt,
  xrightmargin=6pt,
  framexleftmargin=6pt,
  aboveskip=6pt,
  belowskip=6pt,
}
\usepackage{tikz}
\usetikzlibrary{arrows.meta, positioning, fit, backgrounds, calc, shapes.geometric}

\hypersetup{
  colorlinks=true,
  linkcolor=blue!60!black,
  citecolor=blue!60!black,
  urlcolor=blue!60!black,
  pdftitle={Distribird: Literature-Informed Prior Distribution Design for Bayesian Model Calibration},
  pdfauthor={P. P. S\"uli, Gy. Eigner, R. Holl\'os}
}

\mdfdefinestyle{calloutbox}{
  backgroundcolor=gray!10,
  linecolor=gray!50,
  linewidth=0.8pt,
  innertopmargin=8pt,
  innerbottommargin=8pt,
  innerleftmargin=10pt,
  innerrightmargin=10pt,
  skipabove=\baselineskip,
  skipbelow=\baselineskip
}

\mdfdefinestyle{codebox}{
  backgroundcolor=gray!6,
  linecolor=gray!45,
  linewidth=0.6pt,
  innertopmargin=6pt,
  innerbottommargin=6pt,
  innerleftmargin=8pt,
  innerrightmargin=8pt,
  skipabove=6pt,
  skipbelow=6pt
}

\begin{document}

% Title block
\begin{center}
  \includegraphics[width=0.18\textwidth]{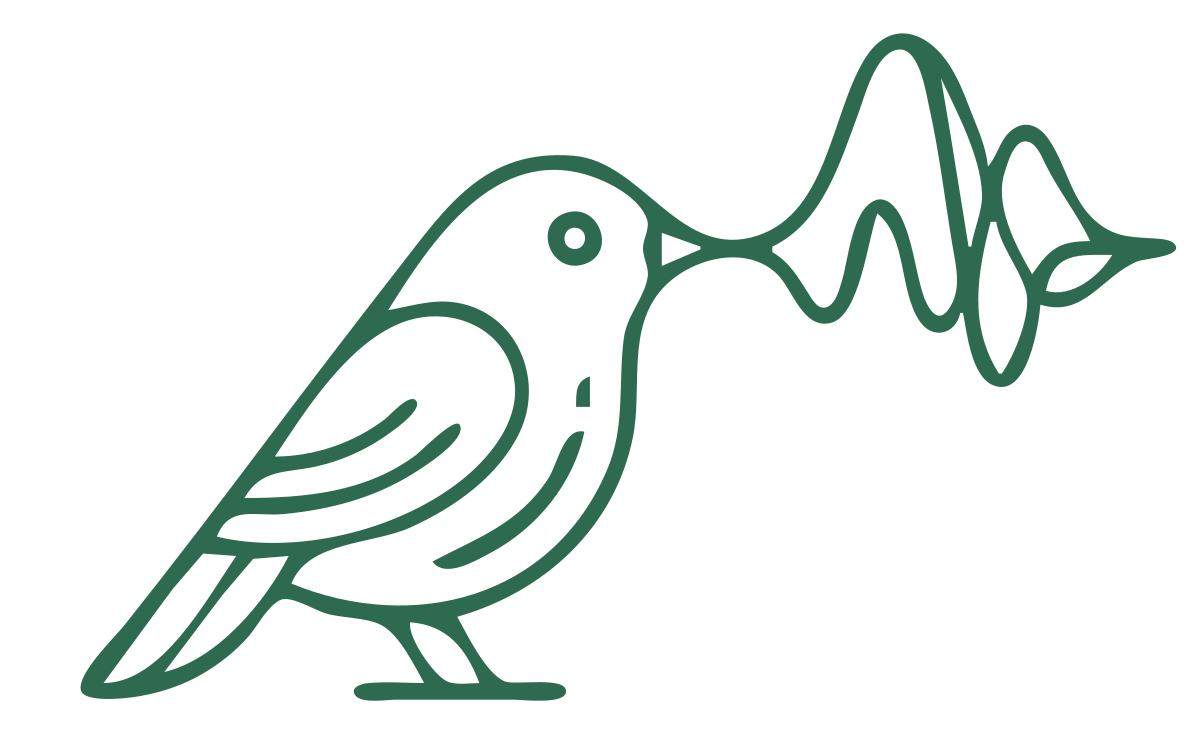}\\[6pt]
  {\LARGE\bfseries Distribird}\\[6pt]
  {\large Literature-Informed Prior Distribution Design for Bayesian Model Calibration}\\[10pt]
  Patrik P.\ S\"uli$^{1,2}$, Gy\"orgy Eigner$^{1,2,3}$, Roland Holl\'os$^{4,5}$\\[6pt]
  {\small
  $^1$ Doctoral School of Applied Informatics and Applied Mathematics, \\Obuda University, Budapest, Hungary\\
  $^2$ Biomatics and Applied Artificial Intelligence Institute, John von Neumann Faculty of Informatics, Obuda University, Budapest, Hungary\\
  $^3$ Physiological Controls Research Center, Obuda University, Budapest, Hungary\\
  $^4$ HUN-REN Centre for Agricultural Research, Brunszvik u. 2., Martonv\'as\'ar 2462, Hungary\\
  $^5$ Czech Academy of Sciences, Global Change Institute, Czech Republic\\
  suli.patrik@uni-obuda.hu, eigner.gyorgy@uni-obuda.hu, hollos.roland@hun-ren.hu
  }\\[4pt]
  2026\\[4pt]

  \textit{Preprint, submitted to arXiv}
\end{center}

\begin{center}
\small \faGlobe\, \href{https://distribird.streamlit.app}{\texttt{distribird.streamlit.app}} \quad | \quad \faGithub\, \href{https://github.com/HUN-REN-AI1Science/Distribird}{\texttt{github.com/HUN-REN-AI1Science/Distribird}}
\end{center}

\vspace{0.25em}

% -----------------------------------------------------------------------
\begin{mdframed}[style=calloutbox]
  \textbf{Abstract}\\[4pt]
  Bayesian calibration of process-based models requires a prior distribution for each model parameter. Despite decades of methodological work, researchers mostly fall back on uniform priors. The main reason is that building informative priors from scientific literature is slow and needs both domain and statistical expertise. We present \textbf{Distribird}, an agentic web application that automates this process. Given a parameter name, physical description, and domain context, Distribird deploys a multi-agent pipeline that searches the literature, extracts and weights reported values by domain relevance, and fits a probability distribution via AIC model selection. When no literature is available, the system falls back to well-defined uninformative alternatives, and clearly reports both the evidence behind and the confidence level of every prior it produces. It is designed for the problems where the models have physically interpretable parameters, where domain knowledge exists in the published literature. We evaluate the tool on 24~parameters across 10 scientific domains comparing open-weight models with a single-prompt LLM baseline. On prior quality the full pipeline \emph{matches} this baseline. Every prior is traced to the specific papers and values from which it was constructed; a built-in validity layer declines to produce priors for out-of-scope requests, whereas the single-prompt baseline returns confident but unfounded priors for them in 11 of 30~model--parameter cases. For scientific use, these properties matter more than a marginal improvement in point-estimate accuracy.
\end{mdframed}

% BullshitBench (out-of-scope detection): see Section~\ref{sec:bullshitbench-design}
% (validity-classification node) and Section~\ref{sec:bullshitbench-eval} (real-LLM
% evaluation across nonsense, theoretical-only, and real parameters).

\vspace{1em}

% -----------------------------------------------------------------------
\section{Introduction}

Process-based models describe complex natural systems (eg. crop growth, water movement through
soil, biogeochemical cycles, disease spread, etc.) using mechanistic equations with parameters that
represent physical, chemical, or biological quantities. Because many of these parameters cannot be measured
directly, they must be inferred from observations by adjusting parameter values until the model
output matches available data. This process is called model calibration, or inverse modeling \citep{Hollos2022}.
Approaches to this parameter-identification problem range from classical optimisation to modern
machine-learning estimators, for instance a neural-network estimator with a differentiable ODE solver
that recovers patient-specific parameters of a tumour-growth model directly from measurements
\citep{Kisbenedek2025}.

Bayesian calibration is arguably the soundest approach to this
problem, on both statistical and philosophical grounds \citep{Gelman2013}. It combines prior knowledge about the parameters with the information
in the observed data, and returns a posterior distribution that captures not only the
most likely parameter values but also the uncertainty around them. This uncertainty quantification
is required for risk-aware decision-making and for complete reporting of model predictions. Explicitly representing
parameter uncertainty is likewise central to robust model-based control of such systems, where the
performance of the controller depends directly on how that uncertainty is captured \citep{Varga2025}.

The Bayesian approach requires the researcher to specify a prior distribution for each parameter
before seeing the data. Every application therefore requires a choice of prior.

The mathematically convenient answer is to use a uniform prior: claim that all parameter
values in a range are equally plausible. This choice is widely adopted in practice
\citep{Wallach2021}. It is also inadequate in most settings. \citet{Pericchi1991} showed
that no single uninformative prior works in every case. Uniform priors in particular are not invariant
under reparameterisation, a well-known limitation: the implied prior changes depending on how the
parameter is written. With a uniform prior the posterior mode coincides exactly with the maximum-likelihood estimate
\citep{Gelman2013}, reducing Bayesian inference to the method it is
intended to improve on. The information lost by ignoring genuine prior knowledge has real
consequences: the sampler converges more slowly, credible intervals are wider than they need to be, and when
data are scarce, posteriors are poorly constrained where an informative prior would have
stabilised them.

Informative priors (priors that capture real knowledge about the parameter) improve all of
these properties; constructing them by reasoning carefully about the parameter, rather than
defaulting to convenience, is a long-standing principle \citep{Gelman1996}. The required knowledge is distributed across the published literature, in primary
measurement reports, review articles, and textbooks; the obstacle is the cost of gathering and combining it. For a model with twenty
parameters, building informative priors from literature can take days of reading, which incentivize the use of uniform priors.
This gap between knowing the problem and having a scalable solution has persisted for more than
thirty years.

Recent Large Language Models (LLMs) that can search databases on their own have created a new
option. An LLM agent can search scientific databases, retrieve relevant papers, extract reported
numerical values, and combine them into a probability distribution automatically, at a
cost that makes the process practical for routine use \citep{Huang2025}. Directly prompting an
LLM for a prior has in fact been shown to yield well-placed, informative distributions quickly
and cheaply \citep{Riegler2025, emodi2025mesh}, which is why we adopt the single-prompt LLM as the baseline
throughout this paper. That same study, however, also identifies important limitations of
directly elicited priors, underscoring the need for caution when they are relied upon in
scientific practice.

LLMs return priors with no traceable basis: as we demonstrate, a model will readily produce a confident, informative prior for a parameter that has no empirical referent, a typographic error, or a software-internal tuning constant. In scientific contexts, where every prior must be defensible and auditable, this is not acceptable. Equally important is the question of attribution: when a language model synthesises findings from the literature without citing its sources, the researchers who produced that empirical work receive no credit, an outcome that undermines the incentive structures on which cumulative science depends.

Our objective is therefore to make literature-based prior construction trustworthy. Concretely, this means three things: every value is traceable to a cited source; out-of-scope requests are declined rather than confabulated; and the entire process runs on open-weight models on the researcher's own hardware, so that no unpublished modelling detail or proprietary data is transmitted to a third-party provider.

We present Distribird, a web application and Python library that implements this pipeline. Distribird targets a specific
problem class: process-based models whose parameters are physically interpretable and reported in an actively published literature; the conditions
are stated in full in the Conditions of Applicability section. This entails an explicit trade-off:
on our benchmark the full pipeline does not produce more accurate priors than a single
well-prompted LLM call, and it is considerably more expensive to run.
It provides provenance, robustness against fabricated evidence, and fully local operation,
the properties that make an automated prior tool suitable for a scientific workflow.

% -----------------------------------------------------------------------
\section{System Design}

Distribird is implemented as a Python~3.10+ package distributed via \href{https://pypi.org/project/distribird/}{PyPI} and exposing two user-facing interfaces: a REST~API built on FastAPI and an interactive Streamlit web application. Both interfaces call the same core pipeline, so the behaviour is identical whichever one is used, and both accept either a single parameter or a batch of parameters, which are processed several at a time in parallel. The system is configured through environment variables (all prefixed \texttt{DISTRIBIRD\_}), so it deploys easily via Docker or a direct installation.

The user provides a \texttt{ParameterInput} object specifying the parameter name, a plain-language physical description, the measurement unit, a domain context string (e.g.\ ``Biome-BGCMuSo maize crop modelling, Central European conditions''), and optional physical constraints (lower bound, upper bound). For example:

\begin{alltt}
  ParameterInput(
      name="TMAX",
      description="Maximum temperature for photosynthesis",
      unit="\textdegree{}C",
      domain_context="Biome-BGCMuSo maize, Central Europe",
      constraints=ConstraintSpec(lower_bound=30.0, upper_bound=50.0)
  )
\end{alltt}

\noindent The system returns a \texttt{PipelineResult} containing the fitted prior distribution, its confidence level, the number of contributing sources, all search queries attempted, the full list of retrieved papers with extracted values, an optional enrichment context, and, when multi-agent deliberation is enabled, the moderator's rationale and any excluded papers. A simplified example:

\begin{verbatim}
  PipelineResult(
      prior=FittedPrior(
          family="truncated_normal",
          params={"mu": 40.5, "sigma": 3.2, "a": 30.0, "b": 50.0},
          confidence="high",
          is_informative=True,
          n_sources=6
      ),
      papers_found=16,
      values_extracted=8,
      search_queries=["maize maximum photosynthesis temperature", ...]
  )
\end{verbatim}

\noindent This complete provenance chain allows the user to audit every step from literature search to final distribution.

\subsection{Multi-agent LangGraph pipeline}

The core of Distribird is a LangGraph\footnote{LangGraph, a library for multi-agent LLM graphs: \url{https://www.langchain.com/langgraph}.} \texttt{StateGraph} that orchestrates a multi-agent pipeline with three feedback loops and a conditional forward path. Unlike a directed acyclic graph, the pipeline allows cycles: when the initial evidence is insufficient, the graph routes back to earlier nodes for another attempt before moving on to synthesis. All nodes read from and write to a shared \texttt{PipelineState} typed dictionary that follows the blackboard pattern: an append-only message log (\texttt{BlackboardMessage}) through which nodes share discoveries, terminology updates, query suggestions, cross-references, and warnings without being wired directly to one another.

\begin{figure}[!htbp]
\centering
\begin{tikzpicture}[
  node distance=0.6cm,
  >={Stealth[length=3pt]},
  mainnode/.style={
    rectangle, rounded corners=3pt, draw=black!70, fill=blue!6,
    minimum width=2.2cm, minimum height=0.65cm,
    font=\footnotesize\sffamily, align=center, inner sep=3pt
  },
  refinenode/.style={
    rectangle, rounded corners=3pt, draw=orange!70!black, fill=orange!10,
    minimum width=2.0cm, minimum height=0.65cm,
    font=\footnotesize\sffamily, align=center, inner sep=3pt
  },
  gatenode/.style={
    diamond, draw=red!60!black, fill=red!6,
    minimum width=1.0cm, minimum height=0.6cm,
    font=\footnotesize\sffamily, align=center, inner sep=1pt,
    aspect=2.2
  },
  validitynode/.style={
    diamond, draw=red!75!black, fill=red!8, line width=0.9pt,
    minimum width=1.2cm, minimum height=0.7cm,
    font=\footnotesize\sffamily, align=center, inner sep=1pt,
    aspect=2.2
  },
  termnode/.style={
    rectangle, rounded corners=3pt, draw=black!50, fill=gray!15,
    minimum width=1.6cm, minimum height=0.5cm,
    font=\footnotesize\sffamily\bfseries, align=center, inner sep=3pt
  },
  looplabel/.style={font=\scriptsize\sffamily\itshape, text=orange!70!black},
  mainedge/.style={draw, ->, thick, black!70},
  loopedge/.style={draw, ->, thick, orange!70!black},
  skipedge/.style={draw, ->, thick, densely dashed, red!70!black},
  condlabel/.style={font=\tiny\sffamily, text=black!50},
  skiplabel/.style={font=\scriptsize\sffamily\bfseries, text=red!80!black},
]
% ============================================================
% MAIN VERTICAL FLOW
% Synthesize runs first; Validity Check is the terminal classifier
% that runs AFTER synthesis and attaches a verdict. The early-skip
% route from Enrich jumps unrecognised parameters straight to that
% terminal node, bypassing the entire search/extract/synthesis machinery.
% A scope-planning node (SearchStrategy) sits between Enrich and QueryGen.
% ============================================================
\node[termnode] (start) {Input};
\node[mainnode, below=of start] (enrich) {Enrich};
\node[mainnode, below=of enrich] (strategy) {Search\\[-1pt]Strategy};
\node[mainnode, below=of strategy] (querygen) {QueryGen};
\node[mainnode, below=of querygen] (search) {Search};
\node[mainnode, below=of search] (relevance) {Relevance\\[-1pt]Judge};
\node[mainnode, below=of relevance] (crossenrich) {CrossEnrich\\[-1pt]{\tiny(snowballing)}};
\node[mainnode, below=of crossenrich] (fulltext) {Fetch\\[-1pt]Fulltext};
\node[mainnode, below=of fulltext] (extract) {Extract};
\node[gatenode, below=0.7cm of extract] (qgate) {Quality\\[-1pt]Gate};
\node[mainnode, below=0.8cm of qgate] (synthesize) {Synthesize};
\node[validitynode, below=0.75cm of synthesize] (validity) {Validity\\[-1pt]Check};
\node[termnode, below=0.7cm of validity] (output) {Output};

% ============================================================
% MAIN FLOW EDGES
% ============================================================
\draw[mainedge] (start) -- (enrich);
\draw[mainedge] (enrich) -- node[condlabel, left] {recognised} (strategy);
\draw[mainedge] (strategy) -- (querygen);

\draw[mainedge] (querygen) -- (search);
\draw[mainedge] (search) -- (relevance);

% --- Conditional: RelevanceJudge -> CrossEnrich OR FetchFulltext ---
% Code: route_after_deliberation() -- if >=2 high-relevance papers -> cross_enrich, else -> fetch_fulltext
\draw[mainedge] (relevance) -- node[condlabel, left] {$\geq$2 relevant} (crossenrich);
\draw[mainedge, densely dashed, black!40]
  ([xshift=0.2cm]relevance.south) -- ++(0,-0.2)
  -| ([xshift=0.55cm]fulltext.east) -- (fulltext.east);
\node[condlabel] at ([xshift=1.1cm]fulltext.east) {otherwise};

\draw[mainedge] (crossenrich) -- (fulltext);
\draw[mainedge] (fulltext) -- (extract);
\draw[mainedge] (extract) -- (qgate);

% --- Quality gate routes to synthesis on the happy path; the terminal
%     validity classifier runs after synthesis and attaches a verdict.
%     Loops A/B/C and the broaden route branch off qgate. ---
\draw[mainedge] (qgate) -- node[condlabel, left] {sufficient} (synthesize);
\draw[mainedge] (synthesize) -- (validity);
\draw[mainedge] (validity) -- (output);

% ============================================================
% EARLY-SKIP ROUTE (highlighted)
% A single red-dashed bypass from Enrich down the far left to the
% terminal Validity node - skipping SearchStrategy, QueryGen, Search,
% RelevanceJudge, CrossEnrich, FetchFulltext, Extract, QualityGate AND
% Synthesize. Code: route_after_enrich() routes unrecognised parameters
% directly to validity_check, which classifies them as LIKELY_INVALID
% with all evidence signals at zero. Validity is terminal (always -> Output),
% so there is no separate invalid-bypass segment.
% ============================================================
\draw[skipedge]
  (enrich.west) -- ++(-5.9,0)
  node[skiplabel, above, midway, yshift=1pt] {EARLY-SKIP}
  node[condlabel, below, midway, text=red!80!black, font=\scriptsize\sffamily] {unrecognised param.}
  |- (validity.west);
% Callout describing what the early-skip avoids
\node[draw=red!70!black, dashed, rounded corners=3pt, fill=red!4,
      text width=3.2cm, font=\scriptsize\sffamily,
      inner sep=4pt, align=left,
      right=2.2cm of validity]
  (skipnote) {%
    \textbf{\textcolor{red!80!black}{Skips entire pipeline}}\\[2pt]
    {\tiny\itshape no search, extract, or synthesis;\\saves 80--95\,\% of runtime}
  };
\draw[draw=red!70!black, dotted, semithick] (validity.east) -- (skipnote.west);

% ============================================================
% MULTI-AGENT SEARCH (annotation to the right of Search node)
% Shows what happens INSIDE search_node
% ============================================================
% Simple callout box listing the agents and moderator
\node[draw=teal!60, rounded corners=4pt, fill=teal!4,
      text width=4.5cm, font=\scriptsize\sffamily,
      inner sep=5pt, align=left,
      right=1.2cm of search, yshift=-0.3cm]
  (agentbox) {%
    \textbf{Parallel source agents:}\\[2pt]
    \textbullet~Semantic Scholar\\
    \textbullet~OpenAlex\\
    \textbullet~Deep Research {\tiny\itshape(opt-in)}\\
    \textbullet~Web Search {\tiny\itshape(opt-in)}\\[3pt]
    \textbf{$\downarrow$ Moderator LLM}\\
    {\tiny deduplicates, selects consensus papers}
  };
\draw[draw, ->, semithick, teal!60!black]
  (search.east) -- (agentbox.west)
  node[condlabel, above, midway, text=teal!60!black] {dispatches};

% ============================================================
% LOOP A: QualityGate -> RefineSearch -> Search
% Code: quality_gate --[0 values]--> refine_search --> search
% ============================================================
\node[refinenode, left=2.0cm of extract] (refinesearch) {Refine\\[-1pt]Search};

\draw[loopedge]
  (qgate.west) -| node[looplabel, above right, pos=0.2] {0 values} (refinesearch.south);
\draw[loopedge]
  (refinesearch.north) |- node[looplabel, above, pos=0.75] {\textbf{Loop A}} (search.west);

% ============================================================
% LOOP B: QualityGate -> BroadenSearch -> QueryGen
% Code: route_after_quality_gate --[too few domain-relevant values]-->
%       broaden_search --> query_gen (regenerate queries at wider breadth).
% Routed down the far left, outside Loop A's channel.
% ============================================================
\node[refinenode, left=2.0cm of querygen] (broaden) {Broaden\\[-1pt]Domain};

\draw[loopedge]
  ([yshift=-0.12cm]qgate.west) -- ++(-4.6,0)
  node[looplabel, below, pos=0.82] {few relevant}
  |- (broaden.west);
\draw[loopedge]
  (broaden.east) -- node[looplabel, above] {\textbf{Loop B}} (querygen.west);

% ============================================================
% LOOP C: QualityGate -> RefineExtraction -> QualityGate
% Code: quality_gate --[low conf.]--> refine_extraction --> quality_gate
% ============================================================
\node[refinenode, right=2.0cm of qgate] (refineextract) {Refine\\[-1pt]Extraction};

\draw[loopedge]
  (qgate.east) -- node[looplabel, above, pos=0.4] {low conf.} (refineextract.west);
\draw[loopedge]
  (refineextract.south) -- ++(0,-0.4) -|
  node[looplabel, below, pos=0.15] {\textbf{Loop C}} ([xshift=0.3cm]qgate.south);

\end{tikzpicture}%
\caption{Distribird LangGraph pipeline with three feedback loops (A: search refinement, B: domain broadening, C: extraction refinement), one cross-enrichment forward path, and an early-skip route (red dashed) from Enrich to Validity~Check for unrecognised parameters. A scope-planning node (SearchStrategy) sets the initial search breadth; Loop~B widens it when on-domain evidence is insufficient. The Search node internally dispatches to parallel source agents whose outputs are reconciled by a Moderator LLM.}
\label{fig:pipeline}
\end{figure}
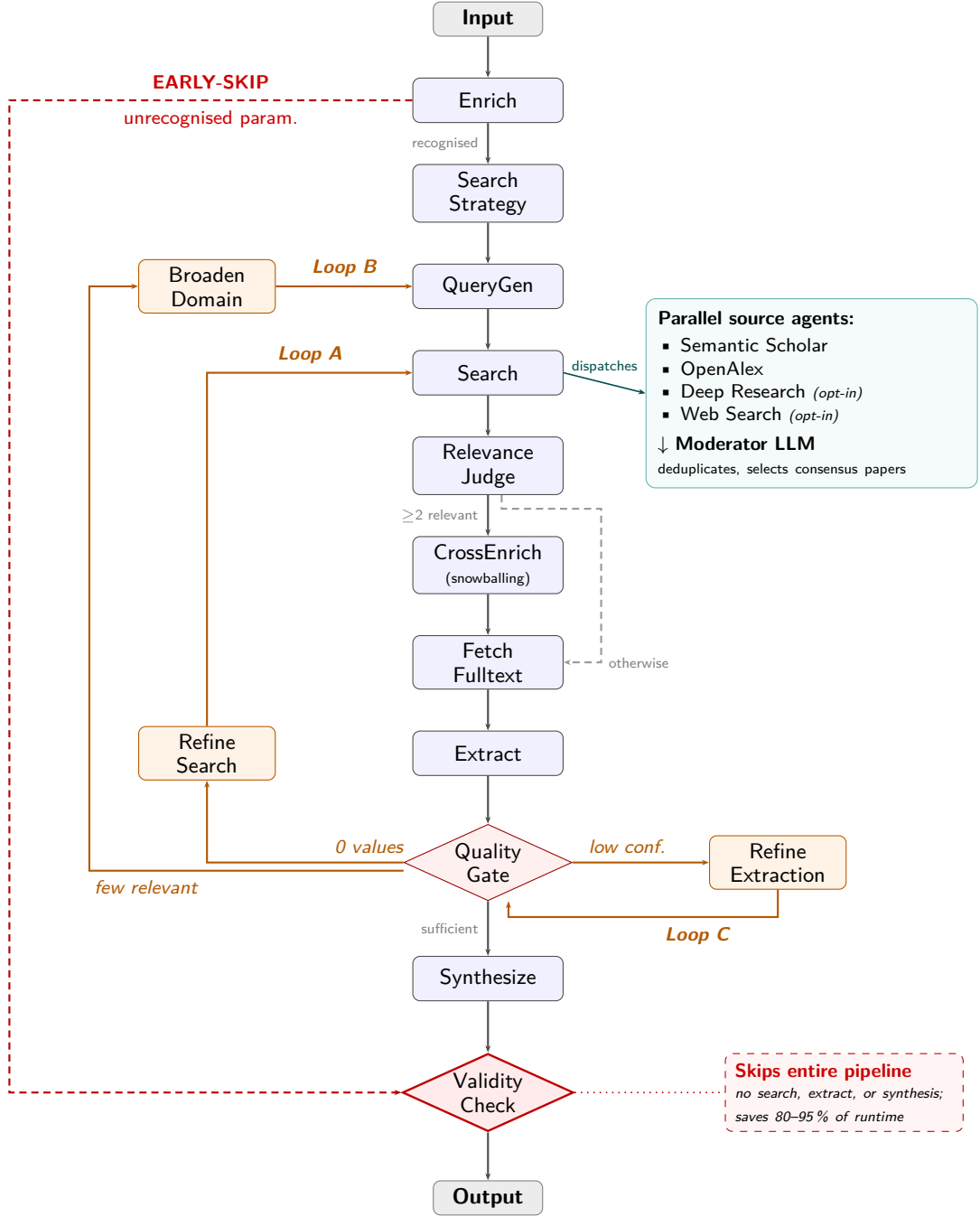

Figure~\ref{fig:pipeline} illustrates the complete graph topology, including three feedback loops, one conditional forward path, and the multi-agent search subsystem. The individual nodes, the three refinement loops, and the routing functions that implement conditional logic are described in the following subsection.

\subsubsection{Pipeline nodes and feedback loops}
\label{sec:feedback}

The pipeline comprises eleven primary processing nodes: Enrich (parameter semantics expansion), SearchStrategy (search-scope planning), QueryGen (search query generation), Search (parallel API queries), RelevanceJudge (paper scoring), CrossEnrich (citation snowballing), FetchFulltext (PDF retrieval and parsing), Extract (numerical value extraction), QualityGate (routing logic), Synthesize (distribution fitting), and ValidityCheck (out-of-scope classification). Three feedback loops and one conditional forward path allow the pipeline to iteratively improve its results when initial evidence is insufficient. Three of these routes are decided at the quality gate, which considers them in priority order (domain broadening first, then search refinement, then extraction refinement) and otherwise proceeds to synthesis; the cross-enrichment forward path is decided separately, right after relevance judgment. The four routes are:

\begin{itemize}
  \item \textbf{Loop~A, Search refinement.} When the quality gate finds zero extracted values but papers were retrieved (indicating that the search found relevant literature but extraction failed to locate numerical data), the pipeline routes to a \texttt{RefineSearch} node. This node analyses the summaries of retrieved papers and generates refined search queries targeting more specific experimental or calibration studies. Control then returns to the Search node. This loop may execute up to two iterations (configurable via \texttt{search\_refinement\_max}).

  \item \textbf{Loop~B, Domain broadening.} When the quality gate finds too few \emph{domain-relevant} values (fewer than \texttt{domain\_broadening\_min\_relevant}, default~2, of the extracted values were judged \textsc{high} or \textsc{medium} relevance to the searched domain) and the search breadth can still be widened, the pipeline routes to a \texttt{BroadenSearch} node. This node advances the current breadth one tier toward broad and returns control to QueryGen, which regenerates queries at the wider scope; papers accumulate across tiers rather than being discarded. This loop may execute up to \texttt{domain\_broadening\_max} times (default~2) and takes priority over the other quality-gate routes.

  \item \textbf{Conditional path, Cross-enrichment.} After relevance judgment, if the maximum iteration number permits and at least two high-relevance papers have been identified, the pipeline routes through CrossEnrich (citation snowballing) before fetching full texts; otherwise, it skips directly to FetchFulltext. This is a conditional forward path, rather than a feedback loop, it does not return control to an earlier node. The cross-enrichment step is executed at most once per pipeline invocation (\texttt{cross\_enrichment\_max\,=\,1}).

  \item \textbf{Loop~C, Extraction refinement.} When the quality gate finds extracted values but none are high-confidence and the coefficient of variation exceeds a predefined threshold (indicating high disagreement among sources), the pipeline routes to a \texttt{RefineExtraction} node that uses web-assisted search to locate additional confirming or disconfirming evidence. Control returns to the quality gate for re-evaluation. This loop may execute once (\texttt{extraction\_refinement\_max\,=\,1}).
\end{itemize}

\noindent An \texttt{IterationBudget} object tracks the number of iterations consumed by each loop and enforces a global cap on total LLM calls (default: 30), guaranteeing termination even when all three loops and the conditional path are activated in the same invocation. The budget is checked at every conditional routing point; when exhausted, the pipeline proceeds directly to synthesis with whatever evidence has been accumulated.

\subsection{Progressive search: scope planning and domain broadening}
\label{sec:progressive}

Because requests vary in how specific they are, the number of requests should be dynamic also. A narrow request (``maximum photosynthesis temperature for \emph{maize} under Central European conditions'') needs a strict first pass that stays close to the exact context, so that unrelated values do not bias the fitted prior. A broad request (``sandstone porosity'') needs a wide first pass, so that a narrow first pass that cannot return sufficient evidence is avoided. Distribird handles both with two cooperating mechanisms, enabled by default (\texttt{enable\_progressive\_search}).

First, a \textbf{scope-planning node} (\texttt{SearchStrategy}) runs immediately after enrichment. A single LLM call classifies the request's domain-specificity as \textsc{high}, \textsc{medium}, or \textsc{low} and maps it to a starting search breadth (strict, mixed, or broad, respectively). If the call is unavailable or the budget is exhausted, the node falls back to a heuristic: a request carrying an application context or context keywords starts strict, otherwise mixed. The chosen breadth conditions the queries generated downstream but never makes the search narrower than the evidence warrants, because of the second mechanism.

Second, a \textbf{domain-broadening loop} (Loop~B above) widens the breadth after the fact when the first pass returns too little on-target evidence. After extraction and the per-paper relevance assessment, the quality gate counts how many extracted values were judged \textsc{high} or \textsc{medium} relevance to the searched domain. If that count is below \texttt{domain\_broadening\_min\_relevant} (default~2) and the breadth is not already at its widest, the \texttt{BroadenSearch} node advances the breadth one tier and returns to query generation. Papers found at each tier accumulate rather than replacing one another, so broadening only ever adds evidence. Values recovered at a wider tier are admitted, but the relevance cap on confidence still governs how strongly they can influence the final prior. Broadening escalates at most \texttt{domain\_broadening\_max} times (default~2), so the pipeline starts as narrow as the request allows and widens only as far as it must.

\subsection{Validity classification: detecting out-of-scope requests}
\label{sec:bullshitbench-design}

A literature-grounded prior only makes sense when the parameter is something the literature actually reports. Two kinds of request fall outside that scope: fabricated names (typos, pseudoscience, placeholders) and real but non-empirical quantities (calibration weights, latent covariances, software-specific tuning factors). In both cases the synthesizer would fall back, without notification to a wide uninformative prior, which a subsequent user could interpret as evidence-based. To prevent this, every request is classified as \textsc{Valid}, \textsc{Suspicious}, \textsc{Likely\_Invalid}, or \textsc{Unknown} (Table~\ref{tab:verdicts}).

\begin{table}[H]
\centering
\caption{The four validity verdicts and the conditions under which each fires.}
\label{tab:verdicts}
\small
\begin{tabularx}{\linewidth}{@{}l>{\raggedright\arraybackslash}X@{}}
  \toprule
  \textbf{Verdict} & \textbf{Conditions} \\
  \midrule
  \textcolor{green!50!black}{\faCheckCircle}~\textsc{Valid} & Recognised parameter; informative prior at medium or high confidence; at least two extracted values. \\[3pt]
  \textcolor{orange!80!black}{\faExclamationTriangle}~\textsc{Suspicious} & Literature exists but no values extractable, or empirical status uncertain. Triggers confirmatory LLM classification call. \\[3pt]
  \textcolor{red!70!black}{\faTimesCircle}~\textsc{Likely\_Invalid} & Unrecognised name (early-skip), or no literature found across the refined queries. \\[3pt]
  \textcolor{black!50}{\faQuestionCircle}~\textsc{Unknown} & Classification disabled, or none of the above rules fired with sufficient evidence. \\
  \bottomrule
\end{tabularx}
\end{table}

Classification uses two gates. The \emph{early-skip} runs right after enrichment: the parameter enrichment prompt is extended with three self-reports (whether the LLM recognises the parameter, the recognition confidence, and whether the quantity is empirically measurable). If the LLM does not recognise the name (at low or zero confidence), the pipeline goes straight to the validity node and skips search, extraction, and synthesis. These are the stages that dominate both runtime (80--95\% in our benchmarks) and LLM token use, since query generation, relevance judging, value extraction, and synthesis are all LLM-driven; skipping them saves both time and API cost on out-of-scope requests. The \emph{terminal classifier} runs at the end of every other request and combines the enrichment self-reports, the number of refined queries, papers, and extracted values, and the prior's confidence into a verdict, using a small set of simple rules. \textsc{Valid} requires recognised terminology, a medium- or high-confidence informative prior, and at least two extracted values. When literature exists but no values could be extracted, or when the empirical status is uncertain, \textsc{Suspicious} fires. \textsc{Likely\_Invalid} activates for the early-skip path and for runs that found nothing at all.

If the heuristics return \textsc{Suspicious}, a single confirmatory LLM call is made: it may revise the verdict to \textsc{Likely\_Invalid} or confirm it as \textsc{Suspicious}, but never overrides a \textsc{Valid} result, and can be disabled for offline use. The final verdict, its reason, and the supporting signals are attached to the pipeline result and shown as a warning in both the Streamlit app and the REST~API response. The whole mechanism costs at most one extra LLM call.

\subsection{Literature search}

The search subsystem queries two academic APIs in parallel. \textit{Semantic Scholar}\footnote{\url{https://www.semanticscholar.org/product/api}} provides citation-graph metadata, abstracts, and open-access PDF links via its Graph~API; \textit{OpenAlex}\footnote{\url{https://openalex.org}} provides an independent index that covers a different set of papers, rebuilding abstracts from its inverted-index form. Both APIs are restricted to open-access papers, so the full text can actually be retrieved.

Two more source agents are available as options. A \textit{deep-research agent} delegates the query to an LLM with web-search capabilities (configurable model, default OpenAI's \texttt{o4-mini-deep-research}) to find papers that the academic APIs may not index; its results are checked against Semantic Scholar before they are included, to guard against fabricated references. A \textit{web search agent} runs a similar checked LLM search using a general-purpose web-search prompt. Both agents contribute \texttt{AgentFinding} objects to the deliberation process alongside the API-based agents.

\subsection{Full-text retrieval and parsing}
\label{sec:fulltext}

Prior-relevant values are rarely reported in abstracts; they appear in calibration tables, methods sections, and appendices. The FetchFulltext node therefore retrieves and reads the complete text of each selected paper, with two design concerns: reaching the content past publisher access restrictions, and preserving its structure once fetched.

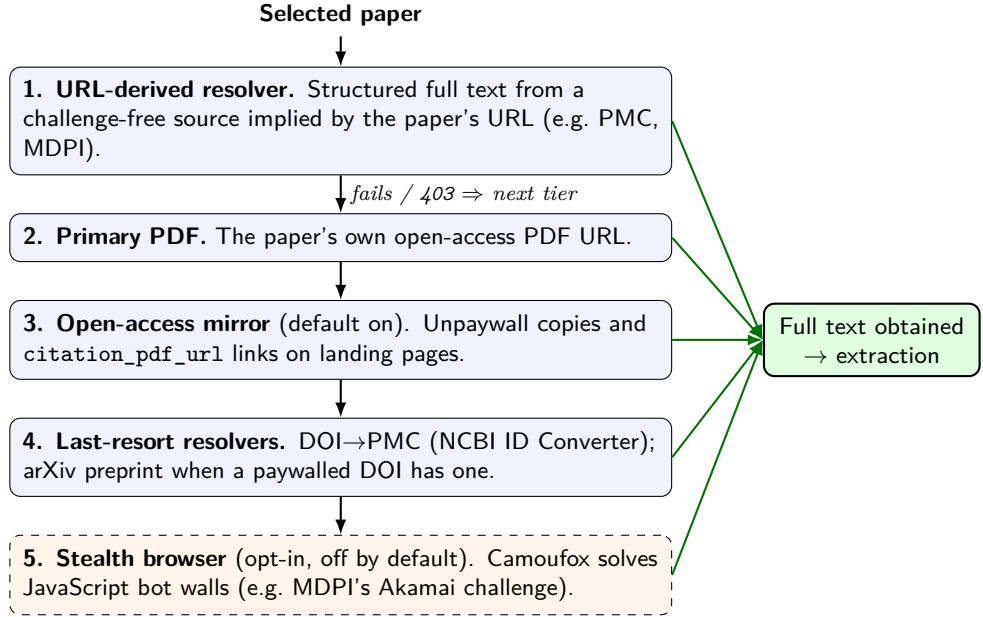
\begin{figure}[t]
\centering
\begin{tikzpicture}[
  font=\footnotesize\sffamily,
  tier/.style={draw, rounded corners, align=left, text width=8.4cm, inner sep=5pt, fill=blue!5},
  optional/.style={draw, dashed, rounded corners, align=left, text width=8.4cm, inner sep=5pt, fill=orange!8},
  win/.style={draw, thick, rounded corners, align=center, text width=2.5cm, inner sep=5pt, fill=green!12},
  fail/.style={-{Latex[length=2mm]}, thick},
  hit/.style={-{Latex[length=2mm]}, green!45!black, thick},
  lbl/.style={font=\scriptsize\itshape}
]
  \node (start) [align=center, font=\footnotesize\sffamily\bfseries] {Selected paper};
  \node (t1) [tier, below=4mm of start] {\textbf{1. URL-derived resolver.} Structured full text from a challenge-free source implied by the paper's URL (e.g.\ PMC, MDPI).};
  \node (t2) [tier, below=5mm of t1] {\textbf{2. Primary PDF.} The paper's own open-access PDF URL.};
  \node (t3) [tier, below=5mm of t2] {\textbf{3. Open-access mirror} (default on). Unpaywall copies and \texttt{citation\_pdf\_url} links on landing pages.};
  \node (t4) [tier, below=5mm of t3] {\textbf{4. Last-resort resolvers.} DOI$\to$PMC (NCBI ID Converter); arXiv preprint when a paywalled DOI has one.};
  \node (t5) [optional, below=5mm of t4] {\textbf{5. Stealth browser} (opt-in, off by default). Camoufox solves JavaScript bot walls (e.g.\ MDPI's Akamai challenge).};
  \foreach \a/\b in {start/t1, t1/t2, t2/t3, t3/t4, t4/t5} \draw[fail] (\a) -- (\b);
  \node[lbl, right=0pt] at ($(t1.south)!0.5!(t2.north)$) {fails / \texttt{403} $\Rightarrow$ next tier};
  \node (win) [win, right=1.2cm of t3] {Full text obtained $\rightarrow$ extraction};
  \foreach \t in {t1,t2,t3,t4,t5} \draw[hit] (\t.east) -- (win.west);
\end{tikzpicture}
\caption{Full-text fetch cascade. Sources are tried top to bottom in increasing order of cost; the first tier that returns usable full text is used(green), remaining tiers are skipped. Tiers 1--4 are pure-HTTP lookups safe on constrained hosts; the dashed tier~5 is a heavyweight, opt-in headless browser used only for JavaScript-based bot-detection challenges.}
\label{fig:fetch}
\end{figure}

A paper's primary open-access PDF URL often fails, most commonly a publisher \texttt{403} returned to non-browser clients. Distribird therefore tries a cascade of sources in increasing order of cost, stopping at the first that returns usable full text (Figure~\ref{fig:fetch}). The early tiers are cheap, pure-HTTP lookups that are safe on constrained hosts such as Streamlit Community Cloud; only the last tier is a heavyweight, opt-in headless stealth browser (Camoufox\footnote{Camoufox, an open-source anti-bot-detection build of Firefox: \url{https://camoufox.com}.}) that is skipped on hosts where it cannot run. Two rules apply throughout the cascade. Whenever a fetched URL returns HTML instead of a PDF, the article text is read from the HTML behind a quality gate that rejects bot-challenge pages and pages containing only the abstract. In addition, an optional institutional HTTP(S) proxy can route every fetch through a subscribing network to reach paywalled papers directly.

By default (\texttt{enable\_markdown\_fulltext}) PDFs are read as Markdown via PyMuPDF4LLM\footnote{PyMuPDF4LLM, a PDF-to-Markdown extractor: \url{https://pypi.org/project/pymupdf4llm/}.} rather than flattened plain text. Tables are preserved as pipe-tables, headings are kept, and page boundaries are marked, so that each reported value reaches the extraction LLM with its row label, column header, and unit association intact rather than flattened into unstructured text. Optical character recognition for scanned pages is available but off by default, because it loads a heavy layout stack that memory-limited hosts cannot run and the fetched open-access papers are usually digital-born with real text layers.

Extraction processes the complete document, not just its opening pages. A paper whose text fits the model's context window is read in a single call; a longer one is split into overlapping ``pages'' sized to the configured window (derived from \texttt{llm\_max\_context\_tokens}), each read separately, and the extracted values are merged and de-duplicated. A safety cap on the number of pages per paper prevents an unusually long PDF from generating hundreds of calls on a small window; when the cap is reached, Methods and Results pages are kept first and the truncation is reported as a warning rather than applied without notification.

\subsection{Prior fitting and confidence hierarchy}
\label{sec:confidence}

The fitting strategy has tiers set by how many values survive extraction and quality-gate filtering; within the top tier, the distribution family is selected from the data by the Akaike Information Criterion (AIC)\citep{Akaike1974}. Table~\ref{tab:fitting-tiers} summarises the four tiers; the rest of this subsection spells each one out.

\begin{table}[H]
\centering
\caption{Tiered prior-fitting strategy. The candidate-family column lists which distributions are considered; the actual choice in the AIC tier depends on the data.}
\label{tab:fitting-tiers}
\small
\begin{tabularx}{\linewidth}{@{}c>{\raggedright\arraybackslash}X>{\raggedright\arraybackslash}Xl@{}}
  \toprule
  \textbf{Values} & \textbf{Method} & \textbf{Candidate family} & \textbf{Confidence} \\
  \midrule
  $\geq 5$ & AIC selection across admissible families & Normal, Truncated Normal, Gamma, Log-Normal, Beta & High \\
  2--4 & Moment matching with $\sigma$ widened $\times 1.5$ and a min-width floor & Truncated Normal & Medium \\
  1 & Centred on the reported value; $\sigma$ from paper's uncertainty, else half the reported value & Truncated Normal & Low \\
  0 & Centred at midpoint of bounds with $\sigma = (u-l)/4$ & Truncated Normal (uninformative) & None \\
  \bottomrule
\end{tabularx}
\end{table}

When five or more values are available, the synthesizer fits five candidate families to the data and selects the family that minimises the AIC ($\mathrm{AIC} = 2k - 2\log\mathcal{L}$), where $k$ is the number of free parameters and $\mathcal{L}$ the maximised likelihood, so that a better fit is rewarded and extra parameters are penalised. All five candidates have $k = 2$ free parameters, so the comparison depends only on the maximised log-likelihood. The candidate families are: \emph{Normal} for unbounded real-valued parameters, \emph{Truncated Normal} for parameters with user-specified physical bounds, \emph{Gamma} for strictly positive quantities with a finite right tail, \emph{Log-Normal} for positive quantities with a heavier right tail, and \emph{Beta} for parameters bounded on a known finite interval. Each family is only entered into the comparison when it is admissible: Gamma and Log-Normal are skipped if any value is $\leq 0$; Beta is skipped unless explicit bounds are supplied and all values lie strictly inside them. Maximum-likelihood estimation uses closed-form mean and standard deviation for Normal and Truncated Normal, and SciPy's \texttt{stats.gamma.fit}, \texttt{stats.lognorm.fit}, and \texttt{stats.beta.fit} for the remaining three families \citep{Virtanen2020}. The winning family is recorded together with its AIC value, and the resulting prior is marked \emph{high confidence}.

When two to four values are available, AIC-based family selection would over-fit. The synthesizer instead applies moment matching against a Truncated Normal: $\mu$ is the (uncertainty-weighted) sample mean, $\sigma$ is the sample standard deviation widened by a factor of~1.5, and a minimum-width floor $\sigma \geq \max(0.05\,|\mu|,\; 0.05\,(u - l))$ prevents an over-narrow prior when all reported values cluster tightly. The prior is marked \emph{medium confidence}.

When exactly one value is available, the synthesizer centres a Truncated Normal on that value, uses the paper's own reported uncertainty as $\sigma$ if present, and otherwise sets $\sigma$ to half the reported value's magnitude (falling back to a small fixed value when the reported value is near zero). The prior is marked \emph{low confidence}.

When zero values are recovered, the synthesizer falls back to a wide Truncated Normal centred at the midpoint of the user-specified bounds with $\sigma = (u - l)/4$, or, if no bounds are supplied, an effectively unbounded Truncated Normal with $\mu = 0$ and $\sigma = 1000$. This is the only tier where the \texttt{is\_informative} flag on \texttt{FittedPrior} is set to \texttt{False}; the prior is marked \emph{none / uninformative}.

\subsubsection{Relevance-aware synthesis}

The tiers above count values but do not ask whether each value actually describes the target domain. For example, a porosity measurement made on carbonate rock is a real, correctly extracted number, but it should not weigh as heavily as one made on the sandstone the user is calibrating. To make this distinction, Distribird performs relevance-aware synthesis by default (\texttt{enable\_fulltext\_relevance}): each value-bearing full-text paper receives a dedicated LLM judgment of how well its study context matches the searched domain and whether it reports usable values, yielding a domain-relevance label of \textsc{high}, \textsc{medium}, or \textsc{low}. This label then changes synthesis in three ways:

\begin{enumerate}[leftmargin=*, itemsep=3pt, topsep=3pt]
  \item \textbf{Selection:} when at least \texttt{relevance\_select\_min\_values} (default~2) values carry \textsc{high} or \textsc{medium} relevance, the distribution is fit from that strongest subset alone. Below that threshold, all in-bounds values are used as a fallback, so a sparse but on-target literature is not discarded.
  \item \textbf{Weighting:} each value's fitting weight is scaled by a moderate relevance weighting factor ($1.0$ for \textsc{high}, $0.6$ for \textsc{medium}, $0.25$ for \textsc{low}), on top of its sample-size weight, so a mislabelled paper is down-weighted rather than excluded without record.
  \item \textbf{Confidence ceiling:} the prior's confidence is capped by the relevance of the values behind it, so that \textsc{high} confidence requires enough genuinely \textsc{high}-relevance evidence and cannot be reached on off-domain values alone.
\end{enumerate}

\noindent When no value carries a relevance label (the feature disabled, or an assessment that could not be made), the confidence cap is not applied and behaviour reduces exactly to the count-only tiers above. The same \textsc{high}/\textsc{medium} relevance count also drives the domain-broadening loop specified in the
progressive-search subsection above.

All fitted distributions are constrained to respect user-specified physical bounds. When a sample size is reported alongside an extracted value, that value is weighted by the square root of its sample size during fitting, so larger studies count for more; a single value's reported uncertainty, when present, is used to set the prior's width. The confidence level is recorded in the \texttt{FittedPrior} object and propagated through all export formats, so that empirically grounded priors remain distinguishable from uninformative fallbacks.

\subsection{Export formats}

Distribird exports priors in three formats designed for direct use in common Bayesian calibration workflows:

\begin{itemize}
  \item \textbf{JSON}, a structured record containing the parameter name, distribution family, fitted parameters, confidence level, informative/uninformative flag, fitting rationale, number of contributing sources, and a citation list with title, DOI, year, and authors for each paper. Batch exports include a version tag and run metadata.

  \item \textbf{Python}, an executable \texttt{scipy.stats} script that defines a frozen distribution object for each parameter \\(e.g.\ \texttt{stats.norm(loc=...,\,scale=...)}) and imports \texttt{numpy} and \texttt{scipy}; a sample-count constant (default: 10\,000) sets how many samples are drawn from the fitted distribution. The generated code is compatible with PyMC, emcee, and custom MCMC samplers.

  \item \textbf{R}, an executable R script using base distribution functions (\texttt{rnorm}, \texttt{rgamma}, \texttt{rlnorm}, \texttt{rbeta}, \texttt{runif}) and the \texttt{truncnorm} package for truncated normal sampling. The script is ready for use with BayesianTools, FME, or custom samplers.
\end{itemize}

\noindent All three formats carry the full provenance chain (from search queries through paper citations to the fitting rationale), so the supporting evidence is exported alongside the fitted parameters.

\subsection{Reproducibility and diagnostics}
\label{sec:reproducibility}

Because every stage that involves an LLM is a potential source of run-to-run variation, Distribird exposes controls that make its output reproducible. An optional integer seed is forwarded to the LLM API's \texttt{seed} field, and the sampling temperature is set per task class (a low ``precise'' temperature for query generation, extraction, relevance judging, and validity; a higher ``creative'' temperature for enrichment and refinement; and a low temperature for the deliberation moderator). Pinning the seed and lowering the temperatures makes repeated runs on the same input converge to the same priors, which is what allows the evaluation to be repeated.

For transparency and debugging, an opt-in debug-trace framework captures a complete structured record of a run, every LLM prompt and response, each search request, each PDF fetch outcome, the extracted values, and the AIC candidates considered during fitting, with sensitive keys redacted. The trace is written to disk, attached to the returned \texttt{PipelineResult}, and can be rendered as a standalone HTML viewer, so any prior can be traced back to the exact evidence and model calls that produced it. Tracing is off by default and behaviour-neutral when disabled.

% -----------------------------------------------------------------------
\section{Evaluation}
\label{sec:exp-eval}

Distribird is built to produce priors that are auditable and reproducible on local hardware, and we evaluate it against exactly those goals. The pipeline delivers three properties that a language model queried directly for a prior cannot. Its priors are \emph{explainable}: every value traces back to a cited paper, so a reviewer can check the evidence behind the distribution. It \emph{declines} to produce priors for fabricated or non-empirical parameters, where a single-prompt model confidently invents one. Finally, it runs entirely on \emph{local open-weight models}, so that no query or document ever leaves the researcher's machine. We quantify each of these, together with the compute cost of running the full pipeline locally.

\subsection{Setup: open-weight models on local hardware}
\label{sec:local-setup}

All runs in this section use open-weight models served locally through \texttt{llama.cpp} as Unsloth Dynamic (UD) GGUF quantisations\footnote{UD quants keep the most error-sensitive layers at higher precision while quantising the rest, giving much smaller files at near-full-precision accuracy: \url{https://docs.unsloth.ai/basics/unsloth-dynamic-2.0-ggufs}.}, so that every language-model call is executed on the machine and no request or unpublished modelling detail is sent to a third-party LLM provider: \textbf{Qwen3.6 27B}\footnote{\url{https://huggingface.co/unsloth/Qwen3.6-27B-GGUF}} and \textbf{Gemma 4 31B}\footnote{\url{https://huggingface.co/unsloth/gemma-4-31b-it-GGUF}} at \texttt{Q8\_K\_XL} (8-bit weights), and \textbf{Mistral Small 4 119B}\footnote{\url{https://huggingface.co/unsloth/Mistral-Small-4-119B-2603-GGUF}} at \texttt{Q4\_K\_XL} (4-bit weights). All three run on a single NVIDIA H100 GPU (80\,GB); the larger Mistral model is quantised to a lower bit-width so that it, too, fits in the H100's memory. Every LLM stage of the pipeline (enrichment, scope planning, query generation, relevance judging, extraction, synthesis, and the validity probe) is served by the same local model. Runs are made reproducible with a fixed seed and the per-task temperatures, and each starts from a cold state with no caches.

As a baseline we use a \emph{naive single-prompt} elicitation: one prompt asks the model directly for a prior distribution (family and parameters), with an explicit option to return an uninformative prior when it lacks knowledge, and with no literature search at all (the exact prompt is given in Appendix~\ref{app:naive-prompt}). We run this baseline on the same three local models and, for reference, on three cloud frontier models (Claude~Opus~4.8, Gemini~3.1~Pro, and GPT-5.5).

The benchmark covers 24~parameters across 12~use cases spanning ten scientific domains, each with a domain-appropriate model and a real, publicly available dataset (Table~\ref{tab:usecases}); ecology contributes three independent predator--prey systems. Each use case contributes two parameters, for example an intrinsic growth rate together with a carrying capacity, or a clearance together with a volume of distribution.

\begin{table}[t]
\centering
\caption{Evaluation use cases and datasets. Each use case contributes two parameters (24 in total). $n$: number of observations.}
\label{tab:usecases}
\footnotesize
\setlength{\tabcolsep}{4pt}
\begin{tabular}{@{}lllr@{}}
  \toprule
  \textbf{Domain / use case} & \textbf{Model / likelihood} & \textbf{Data source} & $n$ \\
  \midrule
  Climate & Linear regr.\ / Normal & AERONET \citep{Holben1998} & 50 \\
  Pharmacokinetics & Bateman eq.\ / Normal & Theoph \citep{Boeckmann1994} & 132 \\
  Hydrology & Bucket / Normal & NRFA stn.\ 39001 & 366 \\
  Structural Eng. & Eigenfreq.\ / Normal & Cable-stayed bridge \citep{Sarmadi2022} & 216 \\
  Epidemiology & SEIR / Neg.\ Binom. & OWID \citep{Mathieu2021} & 60 \\
  Astrophysics & Distance mod.\ / Normal & Pantheon+ \citep{Scolnic2022} & 1701 \\
  Ecology (Isle Royale) & Lotka--Volterra / LogN & \citet{Vucetich2012} & 58 \\
  Ecology (lynx--hare) & Lotka--Volterra / LogN & Hudson Bay pelts \citep{StanLynxHare} & 21 \\
  Ecology (Yellowstone) & Lotka--Volterra / LogN & \citet{Hobbs2024} & 28 \\
  Geophysics & Kozeny--Carman / LogN & USGS \citep{Nelson2003} & 26 \\
  Robotics & Inv.\ dynamics / Normal & KUKA \citep{Meier2016} & 700 \\
  Economics & NK Phillips--Taylor / Normal & FRED \citep{FRED2026} & 60 \\
  \bottomrule
\end{tabular}
\end{table}

\subsection{The cost of a literature-grounded prior}
\label{sec:cost}

Running the full pipeline locally is computationally expensive: it parses the complete text of dozens of papers for every parameter. A single parameter consumes on the order of a million LLM tokens over roughly 80 model calls and 20--40~minutes of time on local hardware (Table~\ref{tab:cost}). Full-text reading and extraction dominate. This cost is a direct consequence of constructing an auditable evidence trail: the same evidence that renders a prior auditable is what makes it expensive to construct. Per-parameter figures (Appendix~\ref{app:local}, Table~\ref{tab:cost-param}) span a wide range: at the lower end of this range, for example, a data-poor parameter costs about 0.4\,M tokens and 12~minutes, whereas at the upper end the Hubble constant, where the literature is vast, reaches 4.7\,M tokens and 96~minutes. This cost is only paid on in-scope requests: the early-skip route settles an unrecognised parameter in seconds.

\begin{table}[t]
\centering
\caption{Per-parameter cost on local hardware, per open-weight model. Tokens (prompt + completion) and time are per parameter (mean; median); calls is the mean number of model calls per parameter.}
\label{tab:cost}
\small
\setlength{\tabcolsep}{6pt}
\begin{tabular}{@{}lccccc@{}}
  \toprule
  \multirow{2}{*}{\textbf{Model}} & \multicolumn{2}{c}{\textbf{Tokens}} & \multicolumn{2}{c}{\textbf{Time}} & \multirow{2}{*}{\textbf{Calls}} \\
  \cmidrule(lr){2-3}\cmidrule(lr){4-5}
   & \textbf{mean} & \textbf{med.} & \textbf{mean} & \textbf{med.} & \\
  \midrule
  Gemma 4 31B          & 1.38\,M & 0.65\,M & 37\,min & 25\,min & 81 \\
  Mistral Small 4 119B & 1.18\,M & 0.77\,M & 21\,min & 18\,min & 81 \\
  Qwen3.6 27B          & 1.21\,M & 0.71\,M & 37\,min & 28\,min & 74 \\
  \bottomrule
\end{tabular}
\end{table}

\subsection{Prior quality: Distribird versus a single-prompt baseline}
\label{sec:prior-quality}

To compare prior \emph{quality} rather than sampling efficiency, we score each prior by its accuracy,
\begin{equation}
\text{accuracy} = 1 - \frac{|\text{prior mode} - \text{data optimum}|}{\text{range}},
\label{eq:accuracy}
\end{equation}
where the data optimum is the maximum-likelihood value on the calibration data (1 = perfect placement, 0 = opposite end of the allowed range). This follows the placement-based evaluation of LLM-suggested priors used by \citet{Riegler2025}, with one modification: we normalise the absolute error by the parameter's allowed range, so that the score is dimensionless and comparable across parameters of widely different scales and units. For each parameter and model we compare the full Distribird prior against the naive single-prompt prior.

Table~\ref{tab:quality} reports the summary, with the full per-parameter breakdown in Appendix~\ref{app:local} (Table~\ref{tab:accuracy-local}). The two approaches place priors about equally well. Across all 24~parameters their mean accuracy is within a few points on every model. Seven of the parameters are benchmark artifacts: the synthetic data force the true value close to the top of the allowed range, (specifically, the data optimum lies at $\geq 85\%$ of the way from the lower to the upper bound of the parameter's allowed range; the seven are listed in Table~\ref{tab:accuracy-local}.) higher than any published value, so no prior consistent with published values can reproduce it (a damping ratio and an incubation period are examples). On the remaining 17 ``fair'' parameters Distribird’s mean accuracy is 7\% points higher on Mistral (83\% vs.\ 76\%) and less than 1\% point lower on Gemma 4 and Qwen.

\begin{table}[H]
\centering
\caption{Prior quality: full Distribird pipeline (informed) vs.\ naive single-prompt baseline, same model and data. Accuracy is mean prior accuracy (100\% = perfect placement), as defined in Equation~\ref{eq:accuracy}; ``fair'' excludes 7 benchmark-artifact parameters.}
\label{tab:quality}
\small
\setlength{\tabcolsep}{6pt}
\begin{tabular}{@{}lcccc@{}}
  \toprule
   & \multicolumn{2}{c}{\textbf{Accuracy (all 24)}} & \multicolumn{2}{c}{\textbf{Accuracy (fair, 17)}} \\
  \cmidrule(lr){2-3} \cmidrule(lr){4-5}
  \textbf{Model} & Informed & Naive & Informed & Naive \\
  \midrule
  Mistral Small 4 119B & 67\% & 62\% & 83\% & 76\% \\
  Gemma 4 31B        & 68\% & 69\% & 84\% & 84\% \\
  Qwen3.6 27B        & 66\% & 67\% & 83\% & 84\% \\
  \bottomrule
\end{tabular}
\end{table}

Beyond comparing means, we test whether the full \emph{distributions} of accuracy scores differ, using a two-sample Kolmogorov--Smirnov test \citep{Massey1951} as implemented in \texttt{scipy.stats.ks\_2samp} over all 72~parameter--model pairs. We run it twice: against the naive baseline on the same local models, and against naive priors from the flagship cloud models (GPT-5.5, Gemini 3.1 Pro, and Claude Opus 4.8). Table~\ref{tab:ks} reports the statistic $D$ and the $p$-value for both, on the full set and on the fair set.

\begin{figure}[H]
\centering
\includegraphics[width=0.82\linewidth]{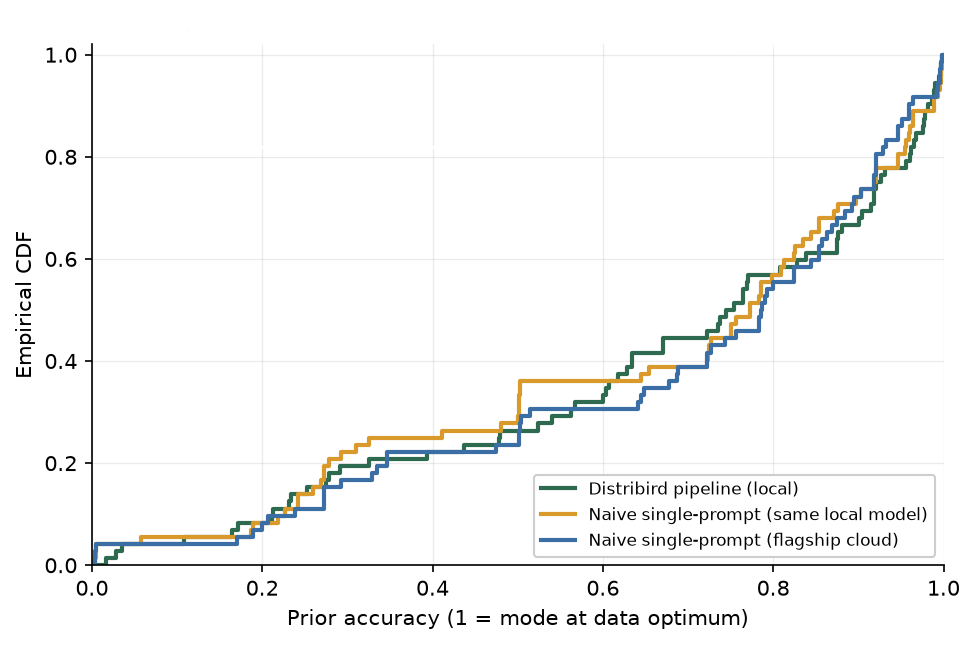}
\caption{Empirical CDFs of prior accuracy (72~parameter--model pairs each) for the Distribird pipeline and the two naive single-prompt baselines.}
\label{fig:ks}
\end{figure}

In every case $D$ is small and the $p$-value is far above the usual $0.05$ threshold, so we cannot reject the hypothesis that the accuracy scores come from one common distribution, and the empirical CDFs are nearly coincident (Figure~\ref{fig:ks}). In other words, the pipeline shows no measurable difference in prior placement from a single well-written prompt, whether that prompt goes to the same local model or to a top cloud model. This is the expected result, since all of them read the same published literature.

\begin{table}[H]
\centering
\caption{Two-sample Kolmogorov--Smirnov statistic $D$ and $p$-value for the Distribird pipeline against each naive baseline. No comparison is significant at $0.05$.}
\label{tab:ks}
\small
\setlength{\tabcolsep}{6pt}
\begin{tabular}{@{}lcccc@{}}
  \toprule
   & \multicolumn{2}{c}{\textbf{All (72)}} & \multicolumn{2}{c}{\textbf{Fair (51)}} \\
  \cmidrule(lr){2-3}\cmidrule(lr){4-5}
  \textbf{Comparison} & $D$ & $p$ & $D$ & $p$ \\
  \midrule
  Pipeline vs.\ naive, same local model & 0.097 & 0.889 & 0.118 & 0.877 \\
  Pipeline vs.\ naive, flagship cloud   & 0.111 & 0.770 & 0.157 & 0.562 \\
  \bottomrule
\end{tabular}
\end{table}

\begin{figure}[H]
\centering
\includegraphics[width=0.82\linewidth]{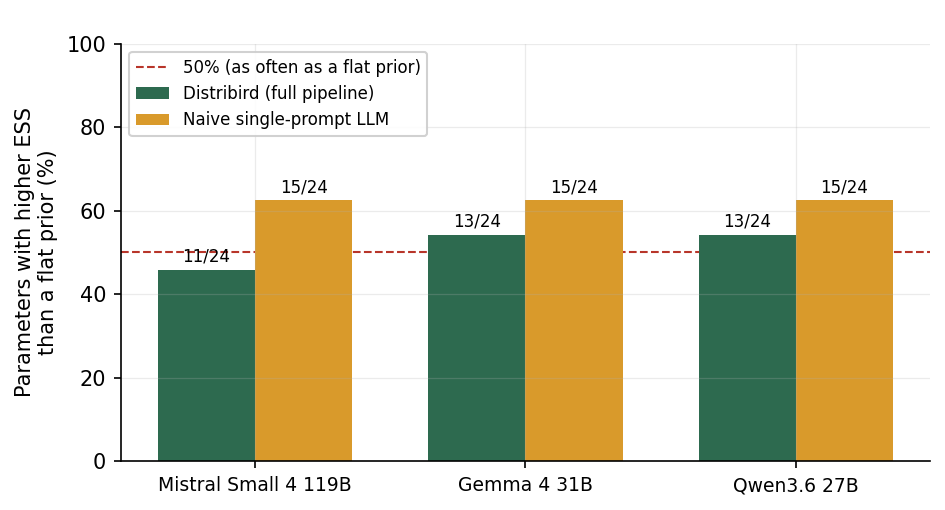}
\caption{Percentage of the 24~parameters whose prior yields a higher MCMC Effective Sample Size (ESS) than a flat prior.}
\label{fig:ess}
\end{figure}

Both priors come from the same language model, so Distribird's literature search mainly shapes the evidence and the width of a prior, not the location of its central tendency. What matters for the sampling step is that both prior types improve sampling efficiency far more often than they degrade it. We measure this with the Effective Sample Size (ESS)\citep{Vehtari2021}, computed from four NUTS chains \citep{Hoffman2014} and compared against a flat prior. Distribird's priors give a higher ESS on 46--54\% of the parameters, and the naive priors on about 62\% (Figure~\ref{fig:ess}). Held-out predictive error improves over a flat prior on about half the use cases, and posterior log-likelihood on almost all of them. The shape of a prior therefore helps even when its centre is slightly off, and for a single point estimate a well-prompted LLM call is already a strong and much cheaper baseline. Distribird, reaches the same quality as the capable LLM's while adding traceable evidence, refusal on out-of-scope requests, and fully local operation, which the rest of this section evaluates.

\subsubsection{Local pipeline versus flagship cloud models}
\label{sec:flagship}
The comparison above holds the model fixed: the pipeline against a naive prompt to the \emph{same} open-weight model. The next comparison is against a naive prompt to a \emph{flagship} cloud model. We scored naive single-prompt priors from GPT-5.5, Gemini 3.1 Pro, and Claude Opus 4.8 on the same 24~parameters, using the same accuracy metric and data optima. These flagship models were run in the single-prompt baseline only, never inside the Distribird pipeline. The comparison is therefore local pipeline versus cloud single-prompt, which is the choice a user actually faces when deciding what to deploy. On the fair set all three flagship baselines score 84\% (Table~\ref{tab:flagship}). This is essentially the same as Distribird on local open-weight models (83--84\%), and the Kolmogorov--Smirnov test confirms it ($D = 0.11$, $p = 0.77$, no significant difference). The local, auditable pipeline therefore places priors about as well as a direct prompt to the highest-scoring cloud models. Its accuracy is statistically indistinguishable from theirs, and it keeps all language-model inference on local hardware, so the researcher's requests and unpublished modelling details never reach a commercial LLM provider.

\begin{table}[H]
\centering
\caption{Naive single-prompt prior accuracy from flagship cloud models, same 24~parameters and metric as Table~\ref{tab:quality}. ``Fair'' excludes the 7 benchmark-artifact parameters. The Distribird row is the mean over the three local open-weight models for reference.}
\label{tab:flagship}
\small
\setlength{\tabcolsep}{8pt}
\begin{tabular}{@{}lcc@{}}
  \toprule
  \textbf{System (prior source)} & \textbf{Accuracy (all 24)} & \textbf{Accuracy (fair, 17)} \\
  \midrule
  Naive single-prompt, GPT-5.5          & 69\% & 84\% \\
  Naive single-prompt, Gemini 3.1 Pro   & 69\% & 84\% \\
  Naive single-prompt, Claude Opus 4.8  & 67\% & 84\% \\
  \midrule
  Distribird, local open-weight (mean)  & 67\% & 83\% \\
  \bottomrule
\end{tabular}
\end{table}

\subsection{Explainability and provenance}
\label{sec:explainability}

What the naive baseline do not provide additional information outside the distribution parameters. . In contrast, Distribird attaches to every prior a complete, machine-readable provenance record: the search queries it issued, the papers it found, the numerical values it extracted with their in-text context, the per-paper domain-relevance judgments (including which values were down-weighted or dropped), the AIC family and score, and a citation list with DOIs.

For example, the COVID-19 incubation-period prior built by Qwen3.6 27B is recorded as an ``AIC-selected lognormal from 85 values (85 high- and medium-relevance values kept, 12 low-relevance dropped)'' drawn from 43~source papers, together with the exact queries issued, the citation-snowballing step that surfaced 18 further key papers, and the citation of every contributing study. A reviewer can follow that chain from the fitted distribution back to the sentence in each paper that produced each value, and discard any source they judge unreliable. This auditability, rather than a lower error, is the reason to prefer the pipeline over a single-prompt model for scientific work.

The difference is clearest where the two approaches disagree. On saturated hydraulic conductivity, the naive single-prompt model had no value it trusted and fell back to an uninformative prior (accuracy~0.50), whereas Distribird on Gemma~4 read 69~papers, kept the 9~values its relevance judge rated \textsc{high} or \textsc{medium}, dropped 6 low-relevance ones, and fit a high-confidence truncated Normal that placed the prior near the data optimum (accuracy~0.91). A reviewer can open those nine papers, check each extracted value, and reverse any inclusion decision; the naive prior provides no comparable means of inspection. The same transparency also exposes Distribird's own limits: for the Isle-Royale moose growth rate, Distribird on Mistral found no usable literature and returned an explicitly uninformative prior, while the naive prompt on that model confidently reported a log-normal whose mass fell largely outside the parameter's bounds (accuracy~0.00). On average across the 24~parameters the two approaches place priors about equally well.

\subsection{Out-of-scope detection}
\label{sec:bullshitbench-eval}

A literature-grounded prior is only meaningful when the parameter is something the literature actually reports. The most consequential failure mode of an automated prior tool is therefore not a slightly-miscentred prior; it is a confident, informative-looking prior generated for a parameter with no empirical basis at all, a typo, a fabricated name, or a software-internal tuning weight, which a downstream user would accept without verification. We test exactly this, applying to prior elicitation the idea of \citet{Gostev2026}'s \emph{BullshitBench}, which measures whether a language model challenges nonsensical prompts instead of answering them confidently.

Our test set is a small, manually constructed proof-of-concept test set of six requests: two fabricated names with no scientific meaning (\texttt{mumblesnort\_\allowbreak factor}, \texttt{fake\_\allowbreak quantum\_\allowbreak correction\_\allowbreak xyz}), three real but non-empirical model-internal quantities (a Biome-BGCMuSo carbon-pool calibration weight \citep{Hollos2022}, a Kalman-filter process-noise covariance entry \citep{Kalman1960}, and a DSSAT root-growth partition factor \citep{Jones2003}), and one real empirical control (\texttt{specific\_leaf\_area}). We run all six through the full Distribird pipeline on the three local models, and through the naive single-prompt baseline on all six models (the three local models and three cloud baselines).

\begin{figure}[H]
\centering
\includegraphics[width=0.86\linewidth]{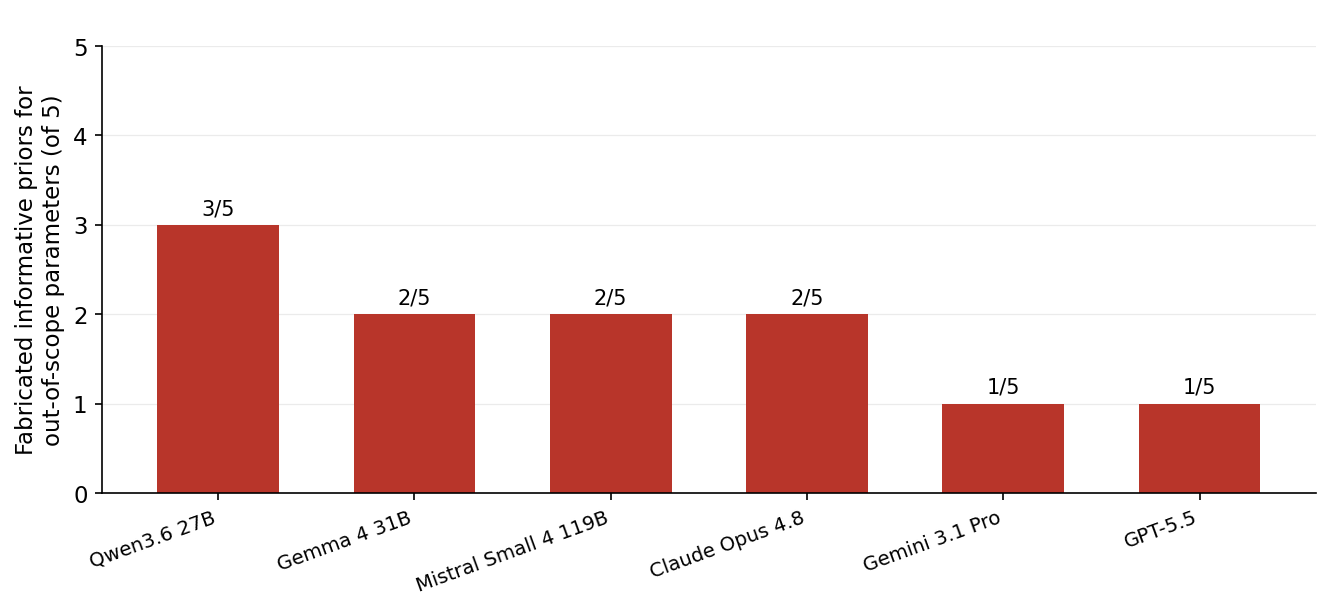}
\caption{Out-of-scope detection. Bars: how often each naive single-prompt baseline invented a confident \emph{informative} prior for a fabricated parameter (out of five). The full Distribird pipeline is not drawn; its per-model verdicts are given in Table~\ref{tab:bullshitbench}.}
\label{fig:bullshit}
\end{figure}

Distribird's validity layer behaves as intended on every model. Across the three local models it flagged all 15 fabricated or non-empirical requests (five per model) as non-valid, \textsc{likely\_invalid} or \textsc{suspicious}, never \textsc{valid} (Table~\ref{tab:bullshitbench}); and it recovered the real control on all three (3/3), returning a literature-backed prior, so it is not simply rejecting all requests. To confirm that the guardrail does not over-reject legitimate requests, we also examined its verdicts on the 24~in-scope benchmark parameters (72~verdicts across the three local models): it never classified a real parameter as \textsc{likely\_invalid} and marked 71\% (51/72) \textsc{valid}; the remaining 29\% received a \textsc{suspicious} verdict, concentrated on the data-poor ecology, robotics, and macroeconomic traits for which few values could be extracted. A \textsc{suspicious} verdict flags a prior for review rather than refusing it, so no genuine parameter is ever discarded. The naive baseline has no such guardrail (Figure~\ref{fig:bullshit}). It returned a confident \emph{informative} prior for the fabricated parameters in 11 of 30 model--parameter cases (the five out-of-scope items across six models) and never once refused; the remaining fabricated cases it answered with an uninformative prior, but never with a warning that the parameter might not be real.

\begin{table}[H]
\centering
\caption{Out-of-scope validity verdicts under the full Distribird pipeline, per local model. No fabricated or non-empirical parameter is ever marked \textsc{valid}; the real control is recovered on all three. \textsc{L\_Inv.}: \textsc{likely\_invalid}; \textsc{Susp.}: \textsc{suspicious}.}
\label{tab:bullshitbench}
\small
\setlength{\tabcolsep}{5pt}
\begin{tabular}{@{}lllll@{}}
  \toprule
  \textbf{Parameter} & \textbf{Category} & \textbf{Qwen} & \textbf{Gemma 4} & \textbf{Mistral} \\
  \midrule
  \texttt{mumblesnort\_factor}          & nonsense       & \textsc{L\_Inv.} & \textsc{L\_Inv.} & \textsc{L\_Inv.} \\
  \texttt{fake\_quantum\_\dots\_xyz}    & nonsense       & \textsc{L\_Inv.} & \textsc{L\_Inv.} & \textsc{L\_Inv.} \\
  \texttt{biome\_bgcmuso\_\dots\_v3}    & non-empirical  & \textsc{L\_Inv.} & \textsc{L\_Inv.} & \textsc{Susp.}   \\
  \texttt{kalman\_\dots\_q11}           & non-empirical  & \textsc{Susp.}   & \textsc{Susp.}   & \textsc{Susp.}   \\
  \texttt{dssat\_\dots\_v45}            & non-empirical  & \textsc{Susp.}   & \textsc{Susp.}   & \textsc{Susp.}   \\
  \texttt{specific\_leaf\_area}         & real (control) & \textsc{Valid}   & \textsc{Valid}   & \textsc{Valid}   \\
  \bottomrule
\end{tabular}
\end{table}

Early refusal also reduces compute cost substantially. Fabricated names are caught by the early-skip route right after enrichment, before any search: on the local models these finished in 12~s to 3.6~min using only 2--6\,k tokens. Requests that reach the full pipeline before being flagged \textsc{suspicious} (the non-empirical quantities that look plausible enough to search for) cost 40\,k--1.2\,M tokens and 7--90~minutes. Detecting that a request is out of scope early therefore saves one to two orders of magnitude of local compute per out-of-scope request, on top of preventing a fabricated prior from ever being returned.

% -----------------------------------------------------------------------
\section{Relation to Existing Work}

Prior elicitation has a large methodological literature \citep{OHagan2006,Garthwaite2005}.
Existing approaches generally fall into two groups: expert elicitation protocols, which
set out a formal process for interviewing domain experts, and empirical Bayes methods, which estimate
prior parameters from data. Distribird fills a different niche: automated literature-based
elicitation, where the prior knowledge comes from the published scientific record rather than from a
human expert or a separate dataset.

Alongside growing interest in LLM-assisted scientific workflows \citep{Boiko2023}, several
groups have explored using language models for statistical analysis tasks. We are not aware of
another system that automates literature-based prior construction end to end, combining multi-agent
literature search, per-source relevance weighting, AIC-based distribution fitting, and explicit
confidence and out-of-scope reporting. Related efforts use language models for statistical analysis
or for literature retrieval, but not to synthesise fitted, provenance-tracked priors.

% -----------------------------------------------------------------------
\section{Conclusion}
\label{sec:conclusion}

In Bayesian calibration of process-based models, selecting appropriate prior distributions is a crucial but time-consuming task that requires synthesizing evidence from the scientific literature and applying domain-expert judgment. Much of the evidence needed to construct informative priors is already documented in scientific publications. Methods for systematically retrieving and synthesizing this evidence could therefore make prior elicitation more efficient. Distribird is an AI-augmented prior-synthesis tool that extracts relevant evidence from the literature, fits candidate probability distributions using the method of moments, and evaluates their relative support using the Akaike Information Criterion (AIC). It produces source-traceable candidate priors for expert review and subsequent use in calibration workflows, or reports when the available evidence is insufficient for reliable distribution fitting.

The evaluation delineates what the pipeline provides. On our benchmark of 24~parameters across
ten domains, the full Distribird pipeline does not produce more accurate priors than a single,
well-prompted call to the same language model, and it is substantially more expensive to run. Its
contribution is trustworthiness rather than accuracy: every prior is traceable to the cited
literature from which it was built, the system declines to produce priors for parameters with no
empirical basis (flagging every fabricated test case, whereas a single-prompt model returns
confident but unfounded priors), and it delivers both on open-weight models running entirely on
local hardware.

The tool is open-source and available from PyPI and as a Docker image. For informal use, a single
LLM call is a cheaper and comparably accurate choice. For scientific use, where a prior must be
defensible, auditable, and robust against silently fabricated evidence, and where transmitting
requests and unpublished model details to a commercial LLM provider may be unacceptable, these are
the properties Distribird provides.

\section{Conditions of Applicability}

Distribird produces its best results under the following conditions, which define the class of
problems the tool is designed for:

\begin{itemize}
  \item \textbf{Physically interpretable parameters:} The parameter has a physical or biological
    interpretation that is discussed in the scientific literature, for example, a maximum
    photosynthesis rate, a root depth, a transmission coefficient, or a thermal threshold.

  \item \textbf{Literature-active domain:} The domain has an active publishing community, so that
    relevant papers can be found by automated search.

  \item \textbf{Univariate distributions:} The parameter can be described by a standard univariate
    probability distribution (Normal, truncated Normal, Beta, Gamma, Log-Normal). Parameters with
    complex multimodal behaviour are outside the current scope.
  \item \textbf{Known physical bounds:} Physical bounds on the parameter are known or can be
    reasoned about by the user, so that the fitted distribution can be constrained appropriately.
\end{itemize}

When these conditions are not fully met, Distribird degrades gracefully: it states the
evidence behind its output clearly, flags low-confidence results, and falls back to
well-defined uninformative alternatives rather than producing false confidence.

\begin{mdframed}[style=calloutbox]
  \textbf{What Distribird is not designed for}\\[4pt]
  Distribird is not designed for parameters without physical interpretation (e.g.\ neural network
  weights), purely empirical tuning coefficients with no literature base, or parameters whose
  behaviour is fundamentally multivariate. For these cases, other approaches (prior predictive
  checks, expert elicitation, or sensitivity analysis) remain more appropriate. To prevent
  undetected misuse, the pipeline classifies each request and explicitly flags out-of-scope inputs.
\end{mdframed}

% -----------------------------------------------------------------------
\section*{Acknowledgments}

We gratefully acknowledge the support of the HUN-REN AI Service Center, which provides critical resources and infrastructure in the form of inspiration, education, consultation, and technology to the research community. Its mission to address researchers' needs in AI and to enable world-class results in scientific inquiry has been instrumental in advancing our work. Gy.\ Eigner was supported by the Distinguished Program of Obuda University.

% -----------------------------------------------------------------------
\bibliographystyle{abbrvnat}

\bibliography{references}

% -----------------------------------------------------------------------
\newpage
\appendix
\section{Per-parameter results on the local runs}
\label{app:local}
This appendix provides the per-parameter detail underlying the aggregate results of Section~\ref{sec:exp-eval}: the cost of constructing each prior, the priors themselves and their sampling efficiency on each open-weight model, and the per-parameter prior accuracy against the naive baseline.
\subsection{Cost per parameter}
Table~\ref{tab:cost-param} reports the mean cost of building each prior across the three local models. Cost scales with the volume of available literature: data-rich, heavily studied quantities read hundreds of papers and cost several million tokens, whereas data-poor traits are resolved in minutes.
\begin{table}[H]\centering
\caption{Per-parameter cost, averaged over the three local models. Tokens: mean total LLM tokens. Time: minutes on local hardware. Papers / Values: mean papers retrieved and numerical values extracted.}
\label{tab:cost-param}
\scriptsize\setlength{\tabcolsep}{5pt}\renewcommand{\arraystretch}{0.95}
\begin{tabular*}{\linewidth}{@{\extracolsep{\fill}}llcccc@{}}\toprule
\textbf{Parameter} & \textbf{Domain} & \textbf{Tokens} & \textbf{Time} & \textbf{Papers} & \textbf{Values} \\\midrule
cloud droplet $r_{\text{eff}}$ & Climate & 2.02\,M & 43\,min & 106 & 83 \\
aerosol optical depth & Climate & 3.54\,M & 79\,min & 133 & 473 \\
clearance & Pharmacok. & 0.46\,M & 13\,min & 55 & 22 \\
volume of distr. & Pharmacok. & 0.74\,M & 19\,min & 69 & 4 \\
field capacity & Hydrology & 0.97\,M & 28\,min & 100 & 9 \\
sat. hydr. conduct. & Hydrology & 0.90\,M & 23\,min & 81 & 12 \\
Young's modulus & Structural & 0.58\,M & 24\,min & 89 & 29 \\
damping ratio & Structural & 0.37\,M & 18\,min & 65 & 9 \\
incubation period & Epidemiol. & 1.74\,M & 38\,min & 109 & 70 \\
infectious period & Epidemiol. & 1.67\,M & 36\,min & 109 & 21 \\
$H_0$ & Astrophys. & 4.69\,M & 96\,min & 107 & 937 \\
$\Omega_m$ & Astrophys. & 4.34\,M & 79\,min & 101 & 431 \\
moose $r$ & Ecol.\ (Isle) & 0.70\,M & 21\,min & 78 & 4 \\
moose $K$ & Ecol.\ (Isle) & 0.59\,M & 18\,min & 84 & 7 \\
hare $r$ & Ecol.\ (lynx) & 0.57\,M & 16\,min & 57 & 3 \\
hare $K$ & Ecol.\ (lynx) & 0.43\,M & 12\,min & 31 & 3 \\
elk $r$ & Ecol.\ (YNP) & 0.50\,M & 18\,min & 78 & 3 \\
elk $K$ & Ecol.\ (YNP) & 0.68\,M & 20\,min & 89 & 1 \\
permeability & Geophysics & 0.52\,M & 16\,min & 61 & 31 \\
porosity & Geophysics & 0.41\,M & 20\,min & 69 & 25 \\
Coulomb friction & Robotics & 0.67\,M & 23\,min & 93 & 3 \\
link inertia & Robotics & 0.69\,M & 30\,min & 100 & 5 \\
price stickiness & Economics & 1.15\,M & 31\,min & 76 & 27 \\
$\phi_\pi$ & Economics & 1.27\,M & 34\,min & 86 & 30 \\
\bottomrule\end{tabular*}\end{table}
\subsection{Constructed priors and sampling efficiency}
Tables~\ref{tab:local-mistral}--\ref{tab:local-qwen} report, for each model, the prior it fitted for every parameter (family with leading parameters; Vals: values extracted; Cf: confidence) together with the resulting Effective Sample Size (ESS) against a flat prior and their ratio. The models frequently select different families and confidence levels for the same parameter, and, as noted in Section~\ref{sec:prior-quality}, the ESS ratio is a secondary diagnostic rather than a measure of prior quality.
\begin{table}[H]\centering
\caption{Constructed prior and sampling efficiency per parameter, \textbf{Mistral Small 4 119B}.\\ ESS$_{\text{inf}}$/ESS$_{\text{flat}}$: Effective Sample Size under the informed vs.\ flat prior; bold ratio: informed win ($>1$).}
\label{tab:local-mistral}
\scriptsize\setlength{\tabcolsep}{4pt}\renewcommand{\arraystretch}{0.95}
\begin{tabular*}{\linewidth}{@{\extracolsep{\fill}}llrc rrr@{}}\toprule
\textbf{Parameter} & \textbf{Prior} & \textbf{Vals} & \textbf{Cf} & \textbf{ESS$_{\text{inf}}$} & \textbf{ESS$_{\text{flat}}$} & \textbf{ratio} \\\midrule
cloud droplet $r_{\text{eff}}$ & LogN(1.88, 0.462) & 38 & H & 11,842 & 9,395 & \textbf{1.26} \\
aerosol optical depth & TN(0.508, 0.368) & 310 & H & 11,862 & 12,675 & 0.94 \\
clearance & LogN(-2.06, 1.54) & 40 & M & 4,802 & 5,070 & 0.95 \\
volume of distr. & TN(0.5, 0.25) & 1 & L & 4,210 & 4,218 & 1.00 \\
field capacity & Beta(1.46, 5.01) & 8 & H & 4,180 & 1,105 & \textbf{3.78} \\
sat. hydr. conduct. & TN(200, 250) & 2 & M & 3,013 & 40 & \textbf{76.22} \\
Young's modulus & TN(30.85, 7.19) & 15 & M & 6,680 & 5,952 & \textbf{1.12} \\
damping ratio & TN(0.04, 0.005) & 1 & L & 6,115 & 6,062 & \textbf{1.01} \\
incubation period & TN(4.28, 0.86) & 20 & H & 6,232 & 7,358 & 0.85 \\
infectious period & TN(7.78, 2.63) & 5 & M & 5,920 & 7,652 & 0.77 \\
$H_0$ & LogN(4.26, 0.0363) & 450 & H & 1,964 & 2,197 & 0.89 \\
$\Omega_m$ & TN(0.303, 0.0589) & 163 & H & 1,775 & 1,918 & 0.93 \\
moose $r$ & TN(0.255, 0.122) & 1 & -- & 542 & 1,067 & 0.51 \\
moose $K$ & TN(3.02, 1.09) & 14 & H & 35 & 8 & \textbf{4.14} \\
hare $r$ & TN(1.17, 0.374) & 3 & L & 5 & 5 & 0.99 \\
hare $K$ & TN(65, 7.5) & 4 & L & 6 & 5 & \textbf{1.22} \\
elk $r$ & TN(0.234, 0.069) & 2 & M & 465 & 342 & \textbf{1.36} \\
elk $K$ & TN(15.5, 7.25) & 2 & -- & 219 & 34 & \textbf{6.36} \\
permeability & Beta(0.936, 48.19) & 33 & H & 2,762 & 270 & \textbf{10.25} \\
porosity & $\Gamma$(2.78, 0.0542) & 15 & M & 2,710 & 273 & \textbf{9.94} \\
Coulomb friction & TN(1, 0.5) & 7 & -- & 4,939 & 6,694 & 0.74 \\
link inertia & TN(0.185, 0.25) & 2 & M & 4,162 & 6,293 & 0.66 \\
price stickiness & Beta(5.11, 10.98) & 8 & L & 6,940 & 4,878 & \textbf{1.42} \\
$\phi_\pi$ & TN(1.5, 0.75) & 5 & L & 4,673 & 5,017 & 0.93 \\
\bottomrule\end{tabular*}\end{table}
\begin{table}[H]\centering
\caption{Constructed prior and sampling efficiency per parameter, \textbf{Gemma 4 31B}.\\ ESS$_{\text{inf}}$/ESS$_{\text{flat}}$: Effective Sample Size under the informed vs.\ flat prior; bold ratio: informed win ($>1$).}
\label{tab:local-gemma}
\scriptsize\setlength{\tabcolsep}{4pt}\renewcommand{\arraystretch}{0.95}
\begin{tabular*}{\linewidth}{@{\extracolsep{\fill}}llrc rrr@{}}\toprule
\textbf{Parameter} & \textbf{Prior} & \textbf{Vals} & \textbf{Cf} & \textbf{ESS$_{\text{inf}}$} & \textbf{ESS$_{\text{flat}}$} & \textbf{ratio} \\\midrule
cloud droplet $r_{\text{eff}}$ & LogN(2.37, 0.352) & 107 & H & 12,534 & 9,395 & \textbf{1.33} \\
aerosol optical depth & TN(0.309, 0.208) & 654 & H & 11,529 & 12,675 & 0.91 \\
clearance & TN(0.0428, 0.0995) & 4 & M & 5,428 & 5,070 & \textbf{1.07} \\
volume of distr. & N(0.458, 0.0436) & 8 & H & 4,441 & 4,218 & \textbf{1.05} \\
field capacity & TN(303, 206) & 4 & M & 2,380 & 1,105 & \textbf{2.15} \\
sat. hydr. conduct. & TN(2927, 1239) & 9 & H & 1,108 & 40 & \textbf{28.04} \\
Young's modulus & LogN(3.5, 0.082) & 18 & M & 7,032 & 5,952 & \textbf{1.18} \\
damping ratio & TN(0.0221, 0.0131) & 16 & M & 6,242 & 6,062 & \textbf{1.03} \\
incubation period & TN(4.58, 0.654) & 26 & H & 6,880 & 7,358 & 0.94 \\
infectious period & LogN(1.87, 0.337) & 11 & H & 7,536 & 7,652 & 0.98 \\
$H_0$ & TN(71.11, 3.15) & 1287 & H & 2,206 & 2,197 & \textbf{1.00} \\
$\Omega_m$ & TN(0.302, 0.0437) & 686 & H & 1,858 & 1,918 & 0.97 \\
moose $r$ & TN(0.306, 0.11) & 7 & H & 2,037 & 1,067 & \textbf{1.91} \\
moose $K$ & TN(2.07, 1.2) & 3 & M & 30 & 8 & \textbf{3.59} \\
hare $r$ & TN(2.36, 0.431) & 2 & M & 5 & 5 & \textbf{1.02} \\
hare $K$ & TN(150, 9.5) & 2 & M & 7 & 5 & \textbf{1.33} \\
elk $r$ & TN(0.255, 0.122) & 0 & -- & 150 & 342 & 0.44 \\
elk $K$ & TN(15.5, 7.25) & 0 & -- & 47 & 34 & \textbf{1.37} \\
permeability & LogN(-1.09, 1.44) & 33 & L & 2,459 & 270 & \textbf{9.13} \\
porosity & TN(0.204, 0.0553) & 11 & H & 2,472 & 273 & \textbf{9.07} \\
Coulomb friction & TN(0.122, 0.0608) & 1 & L & 4,343 & 6,694 & 0.65 \\
link inertia & Beta(0.563, 9.97) & 6 & M & 3,790 & 6,293 & 0.60 \\
price stickiness & TN(3.55, 1.19) & 41 & H & 6,408 & 4,878 & \textbf{1.31} \\
$\phi_\pi$ & Beta(1.66, 6.98) & 61 & H & 5,100 & 5,017 & \textbf{1.02} \\
\bottomrule\end{tabular*}\end{table}
\begin{table}[H]\centering
\caption{Constructed prior and sampling efficiency per parameter, \textbf{Qwen3.6 27B}.\\ ESS$_{\text{inf}}$/ESS$_{\text{flat}}$: Effective Sample Size under the informed vs.\ flat prior; bold ratio: informed win ($>1$).}
\label{tab:local-qwen}
\scriptsize\setlength{\tabcolsep}{4pt}\renewcommand{\arraystretch}{0.95}
\begin{tabular*}{\linewidth}{@{\extracolsep{\fill}}llrc rrr@{}}\toprule
\textbf{Parameter} & \textbf{Prior} & \textbf{Vals} & \textbf{Cf} & \textbf{ESS$_{\text{inf}}$} & \textbf{ESS$_{\text{flat}}$} & \textbf{ratio} \\\midrule
cloud droplet $r_{\text{eff}}$ & LogN(2.33, 0.349) & 57 & M & 12,483 & 9,395 & \textbf{1.33} \\
aerosol optical depth & LogN(-1.5, 1.11) & 338 & H & 14,636 & 12,675 & \textbf{1.15} \\
clearance & LogN(-3, 0.28) & 13 & H & 4,751 & 5,070 & 0.94 \\
volume of distr. & TN(0.434, 0.245) & 4 & M & 4,617 & 4,218 & \textbf{1.09} \\
field capacity & TN(109, 27.5) & 4 & M & 3,265 & 1,105 & \textbf{2.95} \\
sat. hydr. conduct. & Beta(0.448, 3.12) & 5 & M & 2,474 & 40 & \textbf{62.57} \\
Young's modulus & TN(28.5, 8.37) & 36 & M & 7,573 & 5,952 & \textbf{1.27} \\
damping ratio & LogN(-3.91, 0.337) & 5 & M & 6,201 & 6,062 & \textbf{1.02} \\
incubation period & LogN(1.62, 0.304) & 85 & H & 7,948 & 7,358 & \textbf{1.08} \\
infectious period & LogN(2.13, 0.409) & 29 & H & 7,149 & 7,652 & 0.93 \\
$H_0$ & TN(70.91, 2.55) & 829 & H & 1,847 & 2,197 & 0.84 \\
$\Omega_m$ & TN(0.327, 0.0574) & 382 & H & 1,646 & 1,918 & 0.86 \\
moose $r$ & TN(0.334, 0.12) & 4 & M & 1,205 & 1,067 & \textbf{1.13} \\
moose $K$ & TN(2.09, 1.04) & 1 & L & 10 & 8 & \textbf{1.21} \\
hare $r$ & TN(1.5, 0.3) & 1 & L & 5 & 5 & 0.99 \\
hare $K$ & TN(65, 15) & 1 & L & 6 & 5 & \textbf{1.16} \\
elk $r$ & TN(0.195, 0.0675) & 2 & M & 25 & 342 & 0.07 \\
elk $K$ & TN(14.3, 7.15) & 1 & L & 19 & 34 & 0.56 \\
permeability & Beta(0.527, 16.8) & 16 & M & 1,395 & 270 & \textbf{5.18} \\
porosity & TN(0.14, 0.07) & 1 & L & 1,355 & 273 & \textbf{4.97} \\
Coulomb friction & TN(0.47, 0.235) & 1 & L & 8,513 & 6,694 & \textbf{1.27} \\
link inertia & LogN(-1.13, 1.11) & 5 & M & 8,300 & 6,293 & \textbf{1.32} \\
price stickiness & LogN(1.04, 0.509) & 31 & H & 5,054 & 4,878 & \textbf{1.04} \\
$\phi_\pi$ & TN(1.5, 0.204) & 21 & H & 4,157 & 5,017 & 0.83 \\
\bottomrule\end{tabular*}\end{table}
\subsection{Prior accuracy}
Table~\ref{tab:accuracy-local} gives the per-parameter prior accuracy behind the summary in Table~\ref{tab:quality}, for the full pipeline (inf) and the naive single-prompt baseline (nv) on each model.
\begin{table}[H]\centering
\caption{Prior accuracy per parameter (1 = prior mode at the data optimum), full pipeline (inf) vs.\ naive baseline (nv), per model. Rows marked $\dagger$ are benchmark artifacts whose optimum is pinned at $\geq 85\%$ of the way to the upper bound of its range and is unreachable from literature (excluded from the ``fair'' set).}
\label{tab:accuracy-local}
\scriptsize\setlength{\tabcolsep}{5pt}\renewcommand{\arraystretch}{0.95}
\begin{tabular*}{\linewidth}{@{\extracolsep{\fill}}lr rr rr rr@{}}\toprule
 & & \multicolumn{2}{c}{\textbf{Mistral}} & \multicolumn{2}{c}{\textbf{Gemma 4}} & \multicolumn{2}{c}{\textbf{Qwen}} \\
\cmidrule(lr){3-4}\cmidrule(lr){5-6}\cmidrule(lr){7-8}
\textbf{Parameter} & \textbf{opt.} & inf & nv & inf & nv & inf & nv \\\midrule
cloud droplet $r_{\text{eff}}$ & 16.0 & 0.62 & 0.79 & 0.76 & 0.75 & 0.75 & 0.79 \\
aerosol optical depth & 0.44 & 0.98 & 0.90 & 0.96 & 0.87 & 0.87 & 0.88 \\
clearance & 0.039 & 0.99 & 1.00 & 1.00 & 1.00 & 1.00 & 1.00 \\
volume of distr. & 0.48 & 1.00 & 1.00 & 0.99 & 1.00 & 0.99 & 0.99 \\
field capacity & 325 & 0.60 & 0.77 & 0.96 & 0.81 & 0.61 & 0.95 \\
sat. hydr. conduct. & 2500 & 0.54 & 0.50 & 0.91 & 0.50 & 0.56 & 0.50 \\
Young's modulus & 29.6 & 0.97 & 0.99 & 0.90 & 0.96 & 0.97 & 0.99 \\
damping ratio$^\dagger$ & 0.1 & 0.39 & 0.29 & 0.21 & 0.24 & 0.17 & 0.24 \\
incubation period$^\dagger$ & 14.0 & 0.25 & 0.27 & 0.28 & 0.28 & 0.28 & 0.26 \\
infectious period$^\dagger$ & 30.0 & 0.23 & 0.22 & 0.17 & 0.19 & 0.21 & 0.19 \\
$H_0$ & 72.0 & 0.98 & 0.96 & 0.98 & 0.96 & 0.98 & 0.96 \\
$\Omega_m$$^\dagger$ & 0.59 & 0.48 & 0.50 & 0.48 & 0.50 & 0.52 & 0.50 \\
moose $r$ & 0.22 & 0.93 & 0.00 & 0.83 & 0.85 & 0.77 & 0.85 \\
moose $K$ & 0.98 & 0.57 & 0.65 & 0.77 & 0.78 & 0.76 & 0.83 \\
hare $r$ & 1.11 & 0.99 & 0.92 & 0.74 & 0.92 & 0.92 & 0.96 \\
hare $K$ & 80.5 & 0.92 & 0.92 & 0.63 & 0.90 & 0.92 & 0.84 \\
elk $r$ & 0.18 & 0.88 & 0.83 & 0.84 & 0.95 & 0.96 & 0.95 \\
elk $K$ & 25.1 & 0.67 & 0.48 & 0.67 & 0.76 & 0.63 & 0.41 \\
permeability & 2778 & 0.74 & 0.72 & 0.72 & 0.73 & 0.74 & 0.73 \\
porosity$^\dagger$ & 0.36 & 0.33 & 0.33 & 0.60 & 0.65 & 0.44 & 0.31 \\
Coulomb friction & 0.85 & 0.93 & 0.72 & 0.63 & 0.77 & 0.81 & 0.80 \\
link inertia$^\dagger$ & 5.0 & 0.04 & 0.06 & 0.03 & 0.00 & 0.02 & 0.00 \\
price stickiness$^\dagger$ & 12.0 & 0.29 & 0.23 & 0.23 & 0.27 & 0.11 & 0.27 \\
$\phi_\pi$ & 1.0 & 0.88 & 0.75 & 0.90 & 0.81 & 0.88 & 0.82 \\
\bottomrule\end{tabular*}\end{table}

\subsection{Naive single-prompt baseline}
\label{app:naive-prompt}
The naive baseline of Section~\ref{sec:exp-eval} elicits a prior in a single call, with no literature search; braces are per-parameter placeholders. The ``strict'' variant (Section~\ref{sec:bullshitbench-eval}) is identical but drops the abstention bullet (``If you have no useful prior knowledge \ldots'').

\noindent\textbf{System message.}
\begin{lstlisting}[style=promptstyle]
You are a Bayesian statistician with broad scientific knowledge across physics, climate, ecology, pharmacokinetics, epidemiology, structural engineering, geophysics, robotics, and macroeconomics. You answer ONLY in valid JSON - no markdown, no code fences, no preamble. Your output must parse with json.loads().
\end{lstlisting}

\noindent\textbf{User prompt.}
\begin{lstlisting}[style=promptstyle]
Construct a prior probability distribution for the following scientific parameter, drawing exclusively on your training-data knowledge of the relevant literature. You are NOT given any search results - produce your best-effort informed prior from what you already know.

# Parameter
- Name: {name}
- Description: {description}
- Unit: {unit}
- Lower bound: {lower}
- Upper bound: {upper}
- Bounds note: {bounds_note}
- Domain / application context: {domain_context}

# Allowed distribution families (pick ONE)
Use exactly the parameter-key conventions listed; downstream MCMC code reads these keys verbatim.
1. "normal"           - keys: {"mu", "sigma"}
2. "truncated_normal" - keys: {"mu", "sigma", "lower", "upper"}
3. "lognormal"        - keys: {"mu" (log-mean), "sigma" (log-sd)}
4. "gamma"            - keys: {"alpha" (shape), "beta" (rate = 1/scale)}
5. "beta"             - keys: {"alpha", "beta", "lower", "upper"}
6. "uniform"          - keys: {"lower", "upper"}

# Output schema
Return one JSON object with exactly these top-level keys:
{
  "family": "<one of the six families above>",
  "params": { ... family-specific keys as listed above ... },
  "confidence": "<one of: high, medium, low, none>",
  "is_informative": <true if meaningfully tighter than the bounds, else false>,
  "reason": "<one or two sentences: what knowledge you drew on, and why>",
  "typical_range": [<lower>, <upper>],
  "typical_central_value": <your point estimate of the mode or mean>
}

# Guidance
- If you have meaningful prior knowledge of the typical literature values for this parameter, build a prior that concentrates probability mass on that range.
- If you have no useful prior knowledge, return a wide distribution covering the bounds with `is_informative=false` and `confidence="none"`.
- Respect the bounds: chosen family + parameters should put nearly all probability mass inside [lower, upper]. If the family is unbounded (normal, lognormal, gamma) and the bounds are tight, prefer truncated_normal/beta.
- Choose sigma/scale to reflect realistic between-study variation - don't make it so narrow that real data would fall in the prior tails.

Output the JSON object now. No other text.
\end{lstlisting}

\end{document}